\documentclass{article}

\usepackage[final]{neurips_2019}

\usepackage{floatrow}
\newfloatcommand{capbtabbox}{table}[][\FBwidth]

\usepackage[utf8]{inputenc} % allow utf-8 input
\usepackage[T1]{fontenc}    % use 8-bit T1 fonts
\usepackage[colorlinks = true, citecolor = black]{hyperref}       % hyperlinks
\usepackage{url}            % simple URL typesetting
\usepackage{booktabs}       % professional-quality tables
\usepackage{siunitx}
\usepackage{amsfonts}       % blackboard math symbols
\usepackage{nicefrac}       % compact symbols for 1/2, etc.
\usepackage{microtype}      % microtypography

\usepackage{algorithm}
\usepackage{algorithmic}
\usepackage{mathrsfs}

\usepackage{verbatim}
\usepackage{amsmath,amssymb,amsthm,color}
\usepackage{mathtools}
\usepackage{esvect}
\usepackage{bbm}
\usepackage{dsfont}
\usepackage{tikz}
\usepackage{relsize}
\usepackage{adjustbox}
\usetikzlibrary{arrows,decorations.pathmorphing,fit,positioning}
\usepackage{enumitem}
\setlist{leftmargin=*}
\usepackage{units}
\usepackage{microtype}

\usepackage{cleveref}

\usepackage{tikz}
\usepackage{transparent}

\usepackage{times}
\usepackage{graphicx} % more modern
\usepackage{subcaption}
\usepackage{wrapfig}

\theoremstyle{definition}
\newtheorem{defn}{Definition}
\newtheorem{thm}{Theorem}

\newtheorem{lem}{Lemma}

\newcommand{\R}{\mathbb{R}}

\newcommand{\spt}[1]{\text{supp}{\left( #1 \right)}}

\newcommand{\conf}[2]{\mathrm{Conf}_{#1}\!\left(#2 \right)}
\newcommand{\uconf}[2]{\mathrm{UConf}_{#1}\!\left(#2 \right)}

\long\def\comment#1{}

\DeclareMathOperator*{\argmin}{arg\,min}

\definecolor{rouge}{RGB}{255,0,0}
\definecolor{vert}{RGB}{0,128,0}
\definecolor{bleu}{RGB}{0,0,255}

\title{Alleviating Label Switching with Optimal Transport}%A Solution to the Label Switching Problem through Optimal Transport}
\author{
  Pierre Monteiller \\
  ENS Ulm\\
  \texttt{pierre.monteiller@ens.fr} \\
   \And
  Sebastian Claici \\
  MIT CSAIL \& MIT-IBM Watson AI Lab \\
   \texttt{sclaici@mit.edu} \\
   \AND
  Edward Chien \\
  MIT CSAIL \& MIT-IBM Watson AI Lab\\
  \texttt{edchien@mit.edu} \\
   \And
  Farzaneh Mirzazadeh \\
  IBM Research \& MIT-IBM Watson AI Lab\\
   \texttt{farzaneh@ibm.com} \\
  \AND
  Justin Solomon \\
  MIT CSAIL \& MIT-IBM Watson AI Lab\\
  \texttt{jsolomon@mit.edu}\\
  \And
  Mikhail Yurochkin \\
  IBM Research \& MIT-IBM Watson AI Lab\\
  \texttt{mikhail.yurochkin@ibm.com}\\
}

\begin{document}

\maketitle

\begin{abstract}
\emph{Label switching} is a phenomenon arising in mixture model posterior inference that prevents one from meaningfully assessing posterior statistics using standard Monte Carlo procedures. This issue arises due to invariance of the posterior
under actions of a group; for example, permuting the ordering of mixture components has no effect on the likelihood. We propose a resolution to label switching that leverages machinery from optimal transport.  Our algorithm efficiently computes posterior statistics in the quotient space of the symmetry group. We give conditions under which there is a meaningful solution to label switching and demonstrate advantages over alternative approaches on simulated and real data.
\end{abstract}

\section{Introduction}
\label{sec:intro}

Mixture models are powerful tools for understanding multimodal data. In the Bayesian setting, to fit a mixture model to such data, we typically assume a prior number of components and optimize or sample from the posterior distribution over the component parameters. %This posterior distribution depends critically on the prior we have on the data and mixture components. \ed{Why is this last sentence present? It's not like we're messing with the prior at all. We're assuming it's given, right?}
%Mixture models are powerful for combining multiple data streams \justin{really? not clustering?  not sure where ``multiple data streams'' appear} into a single one.  They lie at the heart of many inference algorithms, from expectation maximization to various Monte Carlo samplers (CITE). \justin{previous sentence seems backward; isn't it more like you need EM/MCMC to solve for a mixture model rather than mixture models being at the heart of EM/MCMC?}
If prior components are exchangeable, this
%\justin{``lack'' seems incorrect.  Isn't the problem that the priors \emph{are} exchangeable?}
%\seb{You're right. Brain fart moment}
leads to an identifiability issue known as \emph{label switching}. In particular, permuting the ordering of mixture components does not change the likelihood, since it produces %essentially % <--- justin removed
the same model. The underlying problem is that a group acts on the parameters of the mixture model; posterior probabilities are invariant under the action of the group. %, while the posterior is invariant under group actions.

%The amount of prior information we allow ourselves, however, has a large impact on the performance of mixture models. When we cannot (or will not) put strong priors on our model, Bayesian methods typically prefer exchangeable priors. Unfortunately, this leads to an identifiability problem known as \emph{label switching} (CITE). \justin{I'll admit I couldn't follow this paragraph at all.  Would a Bayesian person be able to follow this?  Seems like a lot of buzzwords (exchangeable prior, identifiability problem, prior information).  Why not just ``tell it like it is'' -- Bayesian posterior can't distinguish between orderings.}

To formalize this intuition, suppose our input is a data set $X$ and a parameter $K$ denoting the number of mixture components. In the most common application, we want to fit a mixture of $K$ Gaussians to the data; our parameter set is $\Theta = \{\theta_1, \ldots, \theta_K\}$ where $\theta_k = \{\mu_k, \Sigma_k, \pi_k\}$ gives the parameters of each component. The likelihood of $x \in X$ conditioned on $\Theta$ is 
%To illustrate this issue, consider a mixture model with parameters $\Theta = \{\theta_1, \ldots, \theta_K\}$ where $\theta_k = \{\mu_k, \pi_k\}$ are respectively the parameter of the prior and the corresponding mixture component weight. \justin{Since it's just an example, maybe just do mixture of Gaussians to avoid clunky phrases like ``respectively the parameter...weight''} Data samples are drawn according to 
%\begin{align*}
    $p(x|\Theta) = \sum_{k=1}^K \pi_k f(x; \mu_k, \Sigma_k),$ 
%\end{align*}
%\justin{wouldn't the posterior be after you apply Bayes' rule to get $p(\Theta|x)$?}
where $f(x;\mu_k,\Sigma_k)$ is the density function of $\mathcal{N}(\mu_k,\Sigma_k)$. Any permutation of the labels $k=1, \ldots, K$ yields the same likelihood. The prior is also permutation invariant. When we compute statistics of the posterior $p(\Theta|x)$, however, this permutation invariance leads to $K!$ symmetric regions in the posterior landscape. 
%The only statistics that can be correctly computed in this context are those invariant to permutations, but the most important statistics---the mean and covariance---are not permutation-invariant. 
%
%Should we permute the labels $k = 1, \ldots, K$ of $\Theta$, then the posterior likelihood remains unchanged. This permutation invariance implies $K!$ symmetric regions in the posterior of $\Theta$. The only sound Bayesian inferences \justin{can ``inference'' be used in this way?} in this context are those that are invariant to the symmetric structure. Means and covariances are not invariant to permutations, which makes it challenging to obtain meaningful statistics from posterior samples.
%
Sampling and inference algorithms behave poorly as the number of modes increases, and this problem is only exacerbated in this context since increasing the number of components in the mixture model leads to a super-exponential increase in the number of modes of the posterior. Previous methods such as the invariant losses of \citet{celeux2000computational} and pivot alignments of \citet{marin2005bayesian} do not identify modes in a principled manner.

To combat this issue, we leverage the theory of optimal transport. In particular, one way to avoid the multimodal nature of the posterior distribution is to replace each sample with its orbit under the action of the symmetry group seen as a distribution over $K!$ points. While this symmetrized distribution is invariant to group actions, we can not average several such distributions using standard Euclidean metrics. We use the notion of a Wasserstein barycenter to calculate a mean in this space, which we can project to a mean in the parameter space via the quotient map. We show conditions under which our optimization can be performed efficiently on the quotient space, thus circumventing the need to store and manipulate orbit distributions with large support.

%To combat this issue, we quotient the parameter space by the action of the symmetry group and work in this space. We leverage the theory of optimal transport, specifically the notion of a Wasserstein barycenter in the space of distributions, to calculate a mean, which naturally allows us to calculate related moments and other posterior statistics. Our key technical contributions relate optimal transport and Wasserstein barycenters on the quotient space to transport and barycenters on the parameter space. When applied to label switching, we obtain an optimization problem for barycenter calculation that can be written as the minimization of the expectation of a simple function, and is thus amenable to stochastic gradient approaches. Our approach enjoys convexity and well-posedness inherited from optimal transport, resolving potential pitfalls of previous efficient algorithms that cope with label switching while maintaining theoretical justification.

%\justin{maybe mention your theory applies to groups acting on variables of a distribution}

\paragraph{Contributions.} We give a practical and simple algorithm to solve the \emph{label switching} problem. To justify our algorithm, we demonstrate that a group-invariant Wasserstein barycenter exists when the distributions being averaged are group-invariant.
%\justin{maybe mention you're extending the Villani paper?} 
We give conditions under which the Wasserstein barycenter can be written as the orbit of a single point, and we explain how failure modes of our algorithm correspond to ill-posed problems. We show that the problem can be cast as computing the expected value of the quotient distribution, and we give an SGD algorithm to solve it.

\section{Related work}\label{sec:labelswitchingrelatedwork}
\textbf{Mixture models.}
Gaussian mixture models are powerful for modeling a wide range of phenomena \citep{mclachlan2019finite}. These models assume that a sample is drawn from one of the latent states (or components), but that the particular component assigned to any given sample is unknown. %Given a number of latent states (or components) in a mixture model, it can be used to estimate the parameters of unknown distributions. The complex nature of such models is simplified by decomposing these models into simpler structures using latent variables. 
%
%  \citep{robert2013monte}
In a Bayesian setup, Markov Chain Monte Carlo can sample from the posterior distribution over the parameters of the mixture model. Hamiltonian Monte Carlo (HMC) has proven particularly successful for this task. Introduced for lattice quantum chromodynamics \citep{duane1987hybrid}, HMC has become a popular option for statistical applications \citep{neal2011mcmc}. Recent high-performance software offers practitioners easy access to HMC and other sampling algorithms \citep{carpenter2017stan}.

\textbf{Label switching.} Label switching arises when we take a Bayesian approach to parameter estimation in mixture models \citep{diebolt1994estimation}. \citet{jasra2005markov} and \citet{papastamoulis2015label} overview the problem. Label switching can happen even when samplers do not explore all $K!$ possible modes, e.g., for Gibbs sampling. Documentation for modern sampling tools mentions that it arises in practice.\footnote{\texttt{\scriptsize{\url{https://mc-stan.org/users/documentation/case-studies/identifying_mixture_models.html}}}} %\href{https://mc-stan.org/users/documentation/case-studies/identifying_mixture_models.html}{label switching in the context of HMC in \texttt{Stan}}}
Label switching can also occur when using parallel HMC, since tools like \texttt{Stan} run multiple chains at once. While a single chain may only explore one mode, several chains are likely to yield different label permutations.

\citet[\S6]{jasra2005markov} mention a few loss functions invariant to the different labelings. Most relevant is the loss proposed by \citet[\S5]{celeux2000computational}.  Beyond our novel theoretical connections to optimal transport, in contrast to their method, our algorithm uses optimal rather than greedy matching to resolve elements of the symmetric group, applies to general groups and quotient manifolds, and uses stochastic gradient descent instead of simulated annealing.
%
%
%An earlier review paper by \citet{jasra2005markov}. This paper presents a somewhat dogmatic Bayesian perspective. Here are some statements made in the paper:
%\begin{itemize}
%    \item Good MCMC algorithm should explore all $K!$ modes as convergence assurance. If it can't explore these modes than it also might not find genuine multimodality in the posterior (i.e. one within same label arrangement). I am not sure the approach we have in mind can handle such multimodality, can it?
%    \item Gibbs sampler seemed to only explore one of the $K!$ modes in their experiments and they strongly recommend against it. Instead they advocate more carefully crafted Metropolis Hastings (MH) based algorithms. In my (limited) experience with MH, it is quite hard to get proposal distribution with reasonable acceptance rate, especially in high dimensions. Gibbs sampler has acceptance rate 1 and seems more popular in the ML literature (but requires local conjugacy, whereas MH conceptually can be applied to any model). More on this later.
%    \item It may help to enforce more informative priors when domain expert knowledge is available, however label switching could still occur.
%    \item Idea of using label invariant loss function for optimization based relabling (see Section 6) seems interesting - maybe we can associate our approach with such loss.
%\end{itemize}
%
%
Somewhat ad-hoc but also related is the pivotal reordering algorithm \citep{marin2005bayesian}, which uses a sample drawn from the distribution as a pivot point to break the symmetry; as we will see in our experiments, a poorly-chosen pivot seriously degrades the performance.
% of this more heuristic approacha poorly-chosen pivot signfi. 
%Data-based relabling in Section 3.6 also might be interesting to relate to. More details to be added. In some of their experiments they used Gibbs sampler to obtain posterior samples, particularly see Section 5.3 and Figures 4 and 5. Gibbs sampler does not explore all $K!$ modes, however it is clear that some label switching is happening and requires correction.

%A bit more \href{https://mc-stan.org/docs/2_18/stan-users-guide/label-switching-problematic-section.html}{here} and \href{https://mc-stan.org/docs/2_18/stan-users-guide/mixture-inference-section.html}{here}. 

%In summary, label switching may also happen when using HMC. Stan runs multiple chains to collect more posterior samples and different chains are very likely to explore different permutations of labels. When mixture components are sufficiently separated, there appears to be no label switching within a single chain --- this raises an interesting question of how to adjust our approach if we may assume that within groups of samples there is no label switching and we need to only find alignment of labels across groups. However when mixture components are not well separated, label switching also occurs within a single chain. In the discussion it is attempted to prevent label switching via structural constraints and non-exchangeable priors. Structural constraints are also discussed in the previous works I mentioned, however they do not generalize to multi-dimensional setting and reasonable non-exchangeable priors are hard to come up with. \\

\textbf{Optimal transport.}
Optimal transport (OT) has seen a surge of interest in learning, from applications in generative models \citep{DBLP:conf/icml/ArjovskyCB17,genevay2018learning}, Bayesian inference \citep{srivastava_wasp:_2015}, and natural language \citep{DBLP:conf/icml/KusnerSKW15,DBLP:conf/emnlp/Alvarez-MelisJ18} to technical underpinnings for optimization methods \citep{DBLP:conf/nips/ChizatB18}. See \citet{solomon2018optimal,peyre2018computational} for discussion of computational OT and \citet{santambrogio_optimal_2015,villani_optimal_2009} for theory.

The Wasserstein distance from optimal transport (\S\ref{sec:preliminaries}) induces a metric on the space of probability distributions from the geometry of the underlying domain. This leads to a notion of a Wasserstein barycenter of several probability distributions \citep{agueh_barycenters_2011}. Scalable algorithms have been proposed for barycenter computation, including methods that exploit entropic regularization \citep{cuturi_fast_2014}, use parallel computing \citep{staib2017parallel}, apply stochastic optimization \citep{DBLP:conf/icml/ClaiciCS18}, and distribute the computation across several machines \citep{DBLP:conf/cdc/UribeDDGN18}.

\section{Optimal Transport under Group Actions}
\label{sec:theory}
Before delving into technical details, we will illustrate our approach with a simple example. Assume we have some data to which we wish to fit a Gaussian mixture model with $K$ components. We can now draw samples from the posterior distribution, and we would like to obtain a point estimate of the mean of the posterior. We draw two samples $\Theta^1 = (\theta_1^1, \ldots, \theta_K^1)$ and $\Theta^2 = (\theta_1^2, \ldots, \theta_K^2)$. We cannot average them due to the ambiguity of label switching; see Figure \ref{fig:schematic}(a) and \S \ref{sec:meanonly} of the supplementary for a simple example. However, we can explicitly encode this multimodality as a uniform distribution over all $K!$ states: 
\begin{align*}
    \frac{1}{K!} \sum_{\sigma\in S_K} \delta_{\sigma \cdot \Theta^1} \quad\text{ and }\quad
   &\frac{1}{K!} \sum_{\sigma\in S_K} \delta_{\sigma \cdot \Theta^2}
\end{align*}
where $S_K$ is the symmetry group on $K$ points that acts by permuting the elements of $\Theta^1$ and $\Theta^2$. These distributions are now invariant to permutations, so we can ask if there exists an average in this space. In this section, we prove that this is possible through the machinery of optimal transport.

We provide theoretical results relevant to optimal transport between measures supported on the quotient space under actions of some group $G$. This theory is fairly general and requires only basic assumptions about the underlying space $X$ and the action of $G$. For each theoretical result, we will use \emph{italics} to highlight key assumptions, since they vary somewhat from proposition to proposition.

%First, a primer on optimal transport, Wasserstein distances, and Wasserstein barycenters. % meh, it's literally the next sentence

\subsection{Preliminaries:  Optimal transport}\label{sec:preliminaries}

Let $(X, d)$ be a \emph{complete} and \emph{separable} metric space. We define the $p$-Wasserstein distance on the space $P(X)$ of probability distributions over $X$ as a minimization over matchings between $\mu$ and $\nu$: 
\begin{align*}
    W_p^p(\mu, \nu) = \inf_{\pi \in \Pi(\mu, \nu)} \int_{X\times X} d(x, y)^p\,\mathrm{d}\pi(x, y).
\end{align*}
Here $\Pi(\mu, \nu)$ is the set of couplings between measures $\mu$ and $\nu$ defined as 
%\begin{align*}
    $\Pi(\mu, \nu) = \{\pi \in P(X\times X)\ |\ \pi(x\times X) = \mu(x), \pi(X\times y) = \nu(y)\}.$
%\end{align*}% justin made this inlined to save space, I know it's ugly but this paper is too long

$W_p$ induces a metric on the set $P_p(X)$ of measures with \emph{finite} $p$-th moments \citep{villani_optimal_2009}. We will focus on $P_2(X)$, %and will understand $P_2(X)$ as a metric space 
endowed with the metric $W_2$. This metric structure allows us to define meaningful statistics for sets of distributions. In particular, a Fr\'echet mean (or Wasserstein barycenter) of a set of distributions $\nu_1, \ldots, \nu_n \in P_2(X)$ is defined as a minimizer 
\begin{equation}\label{eq:finitebarycenter}
    \mu^* = \argmin_{\mu \in P_2(X)} \sum_{i=1}^n \frac{1}{n}W_2^2(\mu, \nu_i).
\end{equation}
%\justin{switched $\mu_i$ to $\nu_i$ to align with story below} \ed{switched ``the'' to ``a'' to acknowledge possibility of multiple Frechet means}
We follow \cite{kim_wasserstein_2017} and generalize this notion slightly, by placing a measure itself on the space $P_2(X)$. We will use $P_2(P_2(X))$ to denote the space of probability measures on $P_2(X)$ that have finite second moments and let $\Omega$ be a member of this set.
%have a measure generating measure $\Omega \in P(P_2(X))$. \justin{I don't know what a ``measure generating measure'' is...} 
%
%\seb{To be extra careful: We might run into trouble if we try to put a distribution over a \emph{subset} of $P_2(X)$ since that subset may not be measurable. \citet{kim_wasserstein_2017} show that the space of absolutely continuous measures $P_{ac}(X)$ is Borel measurable. In our case, we're dealing with the set of one delta distributions. This should also be measurable since you can push forward the volume measure on $X$ via $x \to \delta_x$.} \ed{My understanding of this is that there is a natural measure }
%
Then the following functional will be finite, which generalizes \eqref{eq:finitebarycenter} from finite sums to infinite sets of measures:
\begin{equation}\label{eq:generalbarycenterobjective}
    B(\mu) = \int_{P_2(X)} W_2^2(\mu, \nu)\,\mathrm{d}\Omega(\nu) = \mathbb{E}_{\nu \sim \Omega}\left[W_2^2(\mu, \nu)\right].
\end{equation}
%This expectation exists if the random variable $X(\mu) = W_2^2(\mu, \nu)$ is \emph{bounded} under $\Omega$\justin{would it make sense to say $\nu\sim\Omega$ here?  I had trouble remembering the difference between $\mu$ and $\nu$} \seb{Here where? I use $\nu \sim \Omega$ whenever I talk about the expectation}. This is true if the support of $\Omega$ is \emph{tight}, a notion that we will define shortly.

%\seb{I can't come up with a natural condition here. I think what I want is boundedness of the random variable $X(\mu) = W_2^2(\mu, \nu)$ under $\Omega$. This should be true if the support of $\Omega$ is a tight collection.}

%\seb{What follows is technically new, since \cite{kim_wasserstein_2017} only deal with compact spaces. That said, they do state (without proof) that extensions to non-compact spaces aren't hard (shrug)}\justin{nice!}

In analog to~\eqref{eq:finitebarycenter}, a natural task is to search for a minimizer of the map $\mu \mapsto B(\mu)$. For existence of such a minimizer, we simply require that $\spt{\Omega}$ is tight. 
\begin{defn}[Tightness of measures]
    A collection $\mathcal{C}$ of measures on $X$ is called \emph{tight} if for any $\varepsilon > 0$ there exists a compact set $K \subset X$ such that for all $\mu \in \mathcal{C}$, we have
    %\begin{align*}
        $\mu(K) > 1 - \varepsilon.$
    %\end{align*}
\end{defn}
Here are three examples of tight collections: $P_2(X)$ if $X$ is compact, the set of all Gaussian distributions with means supported on a compact space and of bounded variance, or any set of measures with a uniform bound on second moments (argued in \S \ref{sec:tightness} of the supplementary). This assumption is fairly mild and covers many application scenarios.

Prokhorov's theorem (deferred to the \S \ref{sec:prokhorov}) implies the existence of a barycenter:

\begin{thm}[Existence of minimizers]
\label{thm:existence}
%     The problem
% \begin{equation}
%     \label{eq:barys}
%     \inf_{\mu \in P_2(X)} \mathbb{E}_{\nu\sim \Omega}\left[W_2^2(\mu, \nu)\right].
% \end{equation}
$B(\mu)$ has at least one minimizer in $P_2(X)$ if $\spt{\Omega}$ is tight.
\end{thm}

\subsection{Optimal transport with group invariances}
%Consider now the following setup. 
Let $G$ be a \emph{finite group} that acts by \emph{isometries} on $X$. We define the set of measures invariant under group action $P_2(X)^G = \{\mu \in P_2(X)\ |\ g_{\#}\mu = \mu, \forall g\in G\}$, where the pushforward of $\mu$ by $g$ is defined as  $g_{\#}\mu (B) = \mu(g^{-1}(B))$ for $B$ a measurable set. We are interested in the relation between the space $P_2(X)^G$ and the space of measures on the quotient space $P_2(X/G)$. 
%\justin{should you define $P_2(X)^G$ carefully?} 
%
If all of the measures in the support of $\Omega$ in \eqref{eq:generalbarycenterobjective} are invariant under group action, we can show that there exists a barycenter with the same property:

\begin{lem}
\label{lem:barys}
If $\Omega \in P_2(P_2(X)^G)$ is %a measure 
supported
%\justin{whose support is?}
on the set of group-invariant measures on $X$ and $\spt{\Omega}$ is tight, then there exists a minimizer of $B(\mu)$ in $P_2(X)$ that is invariant under group action.
\end{lem}

\begin{proof}
% If $P$ has finite second moments with respect to $W_2$, we can replace the sum with an expectation over $P$.
Let $\mu\in P_2(X)$ denote the minimizer from Theorem \ref{thm:existence}. Define a new distribution 
%\begin{align*}
    $\mu_G = \frac{1}{|G|} \sum_{g\in G} g_{\#} \mu.$ 
%\end{align*}
We verify that $\mu_G$ has the same cost as $\mu$:
%{\allowdisplaybreaks
\begin{align*}
    &\mathbb{E}_{\nu\sim \Omega}\left[ W_2^2\left(\frac{1}{|G|}\sum_{g\in G} g_\# \mu, \nu\right)\right] \leq \mathbb{E}_{\nu\sim \Omega} \left[\frac{1}{|G|} \sum_{g\in G } W_2^2(g_\# \mu, \nu)\right]\textrm{ by convexity of $\mu \mapsto W_2^2(\mu, \nu)$}\\ 
     &= \mathbb{E}_{\nu\sim \Omega} \left[\frac{1}{|G|} \sum_{g\in G } W_2^2(\mu, (g^{-1})_\# \nu)\right]\textrm{since $g$ acts by isometry}\\
     & =\!\frac{1}{|G|}\sum_{g \in G} \mathbb{E}_{\nu\sim \Omega}\!\left[\!W_2^2(\mu, \nu)\!\right]\!=\!\mathbb{E}_{\nu\sim \Omega}\!\left[\!W_2^2(\mu, \nu)\!\right]\!\textrm{ by linearity of expectation and group invariance of $\nu$.}
\end{align*}    
%}
%where the first line follows via convexity of $\mu \to W_2^2(\mu, \nu)$, the second line by properties of $g$, and the third line by linearity of expectation and group invariance of $\nu$. 
But $\mu$ is a minimizer, so the inequality in line 1 must be an equality.
\end{proof}
\textbf{Remark:} If $X$ is a compact Riemannian manifold and $\Omega$ gives positive weight to the set of absolutely continuous measures, then Theorem 3.1 of \citet{kim_wasserstein_2017} provides uniqueness (and this may be extended to other non-compact cases with suitable decay conditions). However, in our setting, $\Omega$ is supported on samples, measures consisting of delta functions. In this case, a simple counterexample is presented in the supplementary (\S \ref{sec:uniquenesscounter}) which arises in the case where $X$ consists of two points in $\R^2$ and $S_2$ acts to swap the points ($S_K$ is the group of permutations of a finite set of $K$ points). This is accompanied by a study of the case of $K$ points in $\R^d$ (see supplementary \S \ref{sec:meanonly}), relevant to the mixture models where components are evenly weighted and identical with a single mean parameter. Via this study we see that counterexamples seem to require a high degree of symmetry, which is unlikely to happen in applied scenarios, and does not arise empirically in our experiments.

An analogous proof technique can be used to show the following lemma needed later:
\begin{lem}
\label{lem:map}
    If $\nu_1$ and $\nu_2$ are two measures invariant under group action, then there exists an optimal transport plan $\pi\in\Pi(\nu_1, \nu_2)$ that is invariant under the group action $g\cdot \pi(x, y) = \pi(g\cdot x, g\cdot y)$.
\end{lem}

%\begin{prop}[Invariance of Transport Plans]
%\label{prop:plans}
%The transport plans $\pi_i$ between $\mu_i$ and $\nu$ are invariant under the group action $g\cdot \pi_i(x, y) = \pi_i(g\cdot x, g\cdot y)$.
%\end{prop}
%\begin{proof}
%Define $\pi_i^G = \nicefrac{1}{|G|}\sum_{g\in G} g\cdot \pi_i$. The remainder of the proof mirrors that of Proposition %\ref{prop:barys}.
%\end{proof}

The above suggests that we might instead search for barycenters in the quotient space. Consider:  
\begin{lem}[{{\citealt[Lemma 5.36]{lott2009ricci}}}]%[Quotient Map]
\label{lem:quotient}
Let $p : X \to X/G$ be the quotient map. % didn't need it earlier
The map $p_* : P_2(X) \to P_2(X/G)$ restricts to an isometric isomorphism between the set of $P_2(X)^G$ of $G$-invariant elements in $P_2(X)$ and $P_2(X/G)$.
\end{lem}

%The lemma allows us to solve the barycenter problem in $P_2(X/G)$ and pull back the solution to $P_2(X)^G$ by taking the orbit. Lemma \ref{lem:barys} tells us that this procedure is justified and will yield a true barycenter.
%\ed{This lemma tells us that there do exist single orbit barycenters. The previous one tells us there are group-invariant barycenters (assuming minimizers).}

We now introduce additional structure relevant to label switching. Assume that all measures $\nu \sim \Omega$ are the orbits of individual delta distributions, as they are samples of parameter values, i.e., 
%\begin{align*}
    $\nu = \frac{1}{|G|}\sum_{g\in G} \delta_{g\cdot x}$
%\end{align*}
for some $x \in X$. In the simple example of a mixture of two Gaussians from 1D data with means at $\mu_1,\mu_2\in\R$, $\nu$ is of the following form $\nu = \frac{1}{2}\delta_{(\mu_1,\mu_2)}+\frac{1}{2}\delta_{(\mu_2,\mu_1)}$.

Under this assumption and by Lemmas \ref{lem:barys} and \ref{lem:quotient}, 
%We can rephrase our problem under this assumption. Instead of $\Omega\in P_2(P_2(X)^G)$, we can pull back to a distribution $\Omega^*\in P_2(X)$. %Equation \eqref{eq:barys} 
minimization of $B(\mu)$ is equivalent to finding the Wasserstein barycenter of delta distributions on $X/G$. Letting $\Omega_* := p_{*\#} \Omega$, we aim to find:
%\begin{equation}
%    \label{eq:orbit-barys}
%    \inf_{\mu\in P_2(X)} \mathbb{E}_{x\sim \Omega^*}\left[W_2^2\left(\mu, \frac{1}{|G|}\sum_{g\in G} \delta_{g\cdot x}\right)\right]
%\end{equation}
%By Lemma \ref{lem:barys}, we know that a minimizer exists that is invariant under group action. Hence, by Lemma \ref{lem:quotient} we can restrict to the quotient space. 
%Define the push forward of $\Omega$ under $p$ as $\Omega_* = p_*(\Omega)$. In the quotient space, we rewrite \eqref{eq:orbit-barys} as 
\begin{equation}
    \label{eq:quotient-barys}
    \argmin_{\mu\in P_2(X/G)} \mathbb{E}_{\delta_x\sim \Omega_*}\left[W_2^2(\mu, \delta_x)\right].
\end{equation}

From properties of Wasserstein barycenters (\citealt[Equation (2.9)]{carlier_numerical_2015}), the support of $\mu$ lies in the set of solutions to
\begin{equation}\label{eq:frechetquotient}
    \min_{z\in X/G} \mathbb{E}_{\delta_x\sim \Omega_*} \left[d(x, z)^2\right]
\end{equation}
where $d$ is the metric on the quotient space $X/G$ (see e.g.\ \citealt[\S5.5.5]{santambrogio_optimal_2015}). 
%\ed{Do we really know this? They do it only for finitely supported measures on $P(X)$.} \seb{We do, it follows from the multimarginal formulation. Changed the reference}. 
As $\Omega$ has finite second moments, so does $\Omega_*$, giving us existence of the expectation. The existence of minimizers of $z \to \mathbb{E}_{\delta_x\sim \Omega_*} \left[d(x, z)^2\right]$ is established in \S \ref{sec:lemma4} of the supplementary, giving the following lemma:
\begin{lem}
\label{lem:firstMoment}
The map $z \to \mathbb{E}_{\delta_x\sim \Omega_*} \left[d(x, z)^2\right]$ has a minimizer. % didn't need it earlier
\end{lem}
Uniqueness of minimizers is not guaranteed (see \S \ref{sec:uniquenesscounter} of supplementary), but we can rewrite \eqref{eq:quotient-barys} as:
\begin{align*}
    \argmin_{\mu\in P_2(X/G)} \mathbb{E}_{\delta_x\sim \Omega_*}\left[W_2^2(\mu, \delta_x)\right] &= \argmin_{\mu\in P_2(X/G)} \int_{X/G} \int_{X/G} d(x, y)^2 \,\mathrm{d}\mu(y) \,\mathrm{d}\Omega_*(\delta_x)\\
    &= \argmin_{\mu\in P_2(X/G)}\int_{X/G} \int_{X/G} d(x, y)^2 \,\mathrm{d}\Omega_*(\delta_x) \,\mathrm{d}\mu(y).
\end{align*}
By Lemma \ref{lem:firstMoment}, the term $y \to \int_{X/G} d(x, y)^2 \mathrm{d}\Omega_*(\delta_x)$ has a (potentially non-unique) minimizer. Call this function $b(y)$. We are left with
\begin{equation*}
    \argmin_{\mu\in P_2(X/G)}\int_{X/G} b(y) \,\mathrm{d}\mu(y).
\end{equation*}
Any minimizer $y^*$ of $b$ leads to a minimizing distribution $\mu = \delta_{y^*}$, and we can conclude
%but we note that in the case of $X$ being a Riemannian manifold, the Fr\'{e}chet mean of a finite number of points is almost surely unique \citep[Theorem 9]{arnaudon2013medians}. In general, lack of uniqueness seems to require a high amount of symmetry, which is typically not present.
%trickier matter that depends on the set $X$, the group $G$, and the metric. We show later on relevant examples where the minimizer is unique. 
\begin{thm}[Single Orbit Barycenters]
\label{thm:quotientbarys}
    There is a barycenter solution of \eqref{eq:generalbarycenterobjective} that can be written as
    %\begin{align*}
        $\mu = \frac{1}{|G|}\sum_{g\in G} \delta_{g\cdot z^*}.$ 
    %\end{align*}
\end{thm}
%\ed{We don't have uniqueness of the solution because we didn't show that all barycenters are group-invariant. We merely showed that there is a barycenter which is group-invariant.}
Returning to our example of a Gaussian mixture model, we see that this theorem implies there is a barycenter (a mean in distribution space) that has the same form as the symmetrized sample distributions. Any point in the support of the barycenter is an estimate for the mean of the posterior distribution.

%In our example of a mixture of two Gaussians, this theorem states that the barycenter solution is of the same form as $\nu$, i.e. $\frac{1}{2}\delta_{(\mu^*_1,\mu^*_2)}+\frac{1}{2}\delta_{(\mu^*_2,\mu^*_1)}$. The point $z^*= (\mu^*_1,\mu^*_2)$ will be our meaningful statistics for the parameters.

As an aside, we mention that our proofs do not require finite groups. In fact, we prove Theorem \ref{thm:quotientbarys} for compact groups $G$ endowed with a Haar measure in the supplement.

\textbf{To summarize:} Label switching leads to issues when computing posterior statistics because we work in the full space $X$, when we ought to work in the quotient space $X/G$. Theorem \ref{thm:quotientbarys} relates means in $X/G$ to barycenters of measures on $X$ which gives us a principled method for computing statistics backed by a convex problem in the space of measures: take a quotient, find a mean in $X/G$, and then pull the result back to $X$. We will see below in concrete detail that we do not need to explicitly construct and average in $X/G$, but may leverage group invariance of the transport to perform these steps in $X$. 

The crux of this theory is that the Wasserstein barycenter in the setting of Lemma \ref{lem:barys} is a point estimate for the mean of the symmetrized posterior distribution. The results leading to Theorem \ref{thm:quotientbarys} should be understood then as a reduction of the problem of finding an estimate of the mean to that of minimizing a distance function on the quotient space; this latter minimization problem can then be solved via Riemannian gradient descent.
%For example, when we are trying to learn the parameters of a Gaussian mixture model, we can make these algorithms practical by leveraging stochastic gradient approaches, and notions of distance between Gaussian measures.

%\justin{I'd put a short paragraph here summarizing your story in the context of label switching.  Why is this theorem so exciting? Give some intuition or an outline of how this will help you design an algorithm/model.} 

\begin{wraptable}[10]{r}{0pt}

{\scriptsize
\begin{tabular}{|l|l|}
%\vspace{-0.4 in}
\hline
$\mathcal M$ & Riemannian manifold \\
$g_p$ & Inner product at $p\in\mathcal M$\\
$d(p,q)$ & Geodesic distance between $p,q\in\mathcal M$\\
$\mathcal M^K$ & $K$-fold product manifold with product metric\\
$c(p,q)$ & Transport cost, $c(p,q)=\frac{1}{2}d(p,q)^2$\\
$\exp_p,\log_p$ & Exponential, logarithm maps at $p\in\mathcal M$\\
$S_K$ & Symmetric group on $K$ symbols\\
$C_K$ & Cyclic group on $K$ symbols\\
$\mathcal M/G$ & Quotient space of equivalence classes $[p] = \{g\cdot p\ |\ g\in G\}$\\
\hline
\end{tabular}}
\caption{Notation for our algorithm.}\label{table:notation}
\end{wraptable}

\section{Algorithms}
\label{sec:algorithms}

Label switching usually occurs due to symmetries of certain Bayesian models.  Posteriors with the label switching make it difficult to compute meaningful summary statistics, e.g. posterior expectations for the parameters of interest. % As a simple example,
%If we attempt to infer the parameters of our toy  mixture of , % from 1D data with means at $\mu_1,\mu_2\in\R$, 
%due to label switching---which puts equal weight on $(\mu_1,\mu_2)$ as it does on $(\mu_2,\mu_1)$ in the posterior---the expectation over $(\mu_1,\mu_2)$ pairs would be the meaningless pair $(\nicefrac{(\mu_1+\mu_2)}{2},\nicefrac{(\mu_1+\mu_2)}{2})$.

Any attempt to compute posterior statistics in this regime must account for the \emph{orbits} of samples under the symmetry group.  Continuing in the case of expectations, based on the previous section we can extract a meaningful notion of averaging by taking the image of each posterior sample under the symmetry group and computing a barycenter with respect to the Wasserstein metric. %In our simple example, rather than averaging $(\mu_1,\mu_2)$ pairs using the geometry of $\R^2$, we instead take the barycenter of empirical distributions of the form $\frac{1}{2}\delta_{(\mu_1,\mu_2)}+\frac{1}{2}\delta_{(\mu_2,\mu_1)}$ using the geometry of $W_2$ over $P_2(\R^2)$.  
This resolves the ambiguity regarding which points in orbits should match, without symmetry-breaking heuristics like pivoting \citep{marin2005bayesian}.
% (\S\ref{sec:labelswitchingrelatedwork}).

\begin{figure*}
\centering
\def\svgwidth{.9\columnwidth}
\graphicspath{{./figures/}}
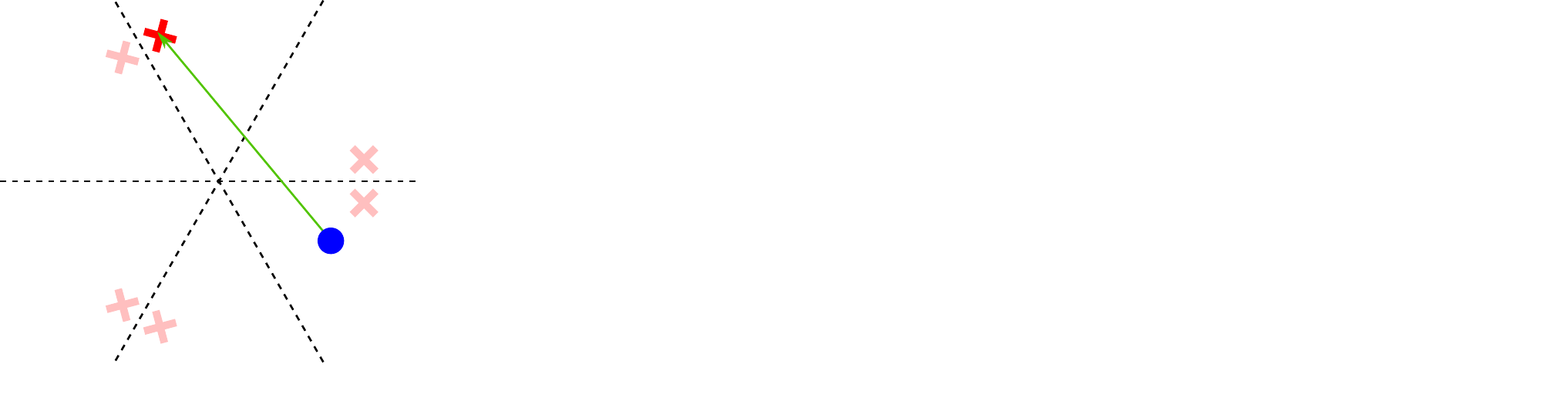
\caption{(a) Suppose we wish to update our estimate of the average (blue) given a new sample (red) from $\Omega$; due to label switching, other points (light shade) have equal likelihood to our sample, causing ambiguity.  (b) Theorem~\ref{thm:quotientbarys} suggests an unambiguous update by constructing $|G|$-point orbits as empirical distributions and doing gradient descent with respect to the Wasserstein metric. (c) This algorithm is equivalent to moving one point, with a careful choice of update functions. This schematic arises for a mean-only model with three means in $\R$ (\S \ref{sec:meanonly} of supplementary); $G = S_3$, with action is generated by reflection over the dashed lines.\vspace{-.15in}}\label{fig:schematic}
\end{figure*}

\begin{wrapfigure}[10]{R}{0.5\textwidth}
\vspace{-.1in}
\begin{minipage}{\linewidth}
\begin{algorithm}[H]
\caption{Riemannian Barycenter of $\Omega$.}
\label{alg:riemannian-mean}
\begin{algorithmic}[1]
    \REQUIRE{Distribution $\Omega$, $\exp$ and $\log$ maps on $\mathcal{M}$}
    \ENSURE{Estimate of the barycenter of $\Omega$}
    \STATE{Initialize the barycenter $p \sim \Omega$}.
    \FOR{$t=1, \ldots$}
        \STATE{Draw $q\sim \Omega$}
        \STATE{$-D_{p}c(p,q) \coloneqq \log_{p}(q)$}
        \STATE{$p\gets \exp_{p}\left(-\frac{1}{t}D_{p}c(p,q)\right)$}
    \ENDFOR
\end{algorithmic}
\end{algorithm}
\end{minipage}
\end{wrapfigure}

In this section, we provide an algorithm for computing the $W_2$ barycenters above, extracting a symmetry-invariant notion of expectation for distributions with label switching.  As input, we are given a sampler from a distribution $\Omega$ over a space $\mathcal M$ subject to label switching, as well as its (finite) symmetry group $G$.  Our goal is to output a barycenter of the form $\frac{1}{|G|}\sum_{g\in G}\delta_{g\cdot x}$ for some $x\in\mathcal M$, using stochastic gradient descent on~\eqref{eq:generalbarycenterobjective}.  Our approach can be interpreted two ways, echoing the derivation of Theorem~\ref{thm:quotientbarys}:
\begin{itemize}
    \item The most direct interpretation, shown in Figure~\ref{fig:schematic}(b), is that we push forward $\Omega$ to a distribution over empirical distributions of the form $\frac{1}{|G|}\sum_{g\in G}\delta_{g\cdot x},$ where $x\sim\Omega$, and then compute the barycenter as a $|G|$-point empirical distribution whose support points move according to stochastic gradient descent, similar to the method by~\citet{DBLP:conf/icml/ClaiciCS18}. 
    \item Since $|G|$ can grow extremely quickly, we argue that this algorithm is \emph{equivalent} to one that moves a single representative $x$, so long as the gradient with respect to $x$ accounts for the objective function; this is illustrated in Figure~\ref{fig:schematic}(c).
\end{itemize}
Although our final algorithm has cosmetic similarity to pivoting and other algorithms that compute a single representative point, the details of our approach show an \emph{equivalence} to a well-posed transport problem.  Moreover, our stochastic gradient algorithm invokes a sampler from $\Omega$ in every iteration, rather than precomputing a finite sample, i.e.  our algorithm deals with samples as they come in, rather than collecting multiple samples, and then trying to cluster or break the symmetry \emph{a posteriori}.

Table~\ref{table:notation} gives a reference for the notation used in this section.  Note the Riemannian gradient of $c(p, q)$ has a particularly simple form: $-D_p c(p, q) = \log_p(q)$ \citep{kim_wasserstein_2017}.%\ed{we should cite something here} \justin{maybe push this to where it's used}

\begin{wrapfigure}[13]{R}[10 pt]{0.5\textwidth}
\vspace{-.17in}
\begin{minipage}{\linewidth}
\begin{algorithm}[H]
    \caption{Barycenter of $\Omega$ on quotient space}\label{alg:barycenters}
    \begin{algorithmic}[1]
    
    \REQUIRE{Distribution $\Omega$, $\exp$ and $\log$ maps on $\mathcal{M}$}
    \ENSURE {Barycenter $[(p_1, \ldots, p_K)]$}
    \STATE{Initialize the barycenter $(p_1, \ldots, p_K) \sim \Omega$}.
    \FOR{$t=1, \ldots$}
        \STATE{Draw $(q_1, \ldots, q_K) \sim \Omega$}
        \STATE{Compute $\sigma$ in \eqref{eq:quotientdist}}
        \FOR{$i=1, \ldots, K$}
            \STATE{$-D_{p_i}c(p_i,q_{\sigma(i)}) \coloneqq \log_{p_i}(q_{\sigma(i)})$}
            \STATE{$p_i\gets \exp_{p_i}\left(-\frac{1}{t}D_{p_i}c(p_i,q_{\sigma(i)})\right)$}
        \ENDFOR
    \ENDFOR
    \end{algorithmic}
\end{algorithm}
\end{minipage}
\end{wrapfigure}

\textbf{Gradient descent on the quotient space.}
For simplicity of exposition, we introduce a few additional assumptions on our problem; our algorithm can generalize to other cases, but these assumptions are the most relevant to the experiments and applications in \S\ref{sec:results}.  
In particular, we assume we are trying to infer a mixture model with $K$ components. The parameters of our model are tuples $(p_1, \ldots, p_K)$, where $p_i \in \mathcal{M}$ for all $i$ and some Riemannian manifold $\mathcal{M}$. We can think of the space of parameters as the product $\mathcal{M}^K$. Typically it is undesirable when two components match exactly in a mixture model, so we additionally excise any tuple $(p_1, \ldots, p_K)$ with any matching elements (together a set of measure zero). Representing parameters in a mixture model can be made through a point process, it is natural to work with the $K$th ordered configuration space of  $\mathcal{M}$ considered in physics and algebraic topology \citep{fadell2001configspaces}:
\[ \conf{K}{\mathcal{M}} := \mathcal{M}^K \big\backslash \{(p_1, \ldots, p_K) \mid p_i = p_j \, \mathrm{for}\, \mathrm{some}\, i \neq j\} \subset \mathcal{M}^K. \]
% ,two  after observing data, $\mathcal M^K$ is the space on which our Bayesian posterior is supported. 

Let $\Omega \in P(\conf{K}{M})$ be the Bayesian posterior distribution restricted to $\conf{K}{M}$ (assuming the posterior $P(\mathcal{M}^K)$ is absolutely continuous with respect to the volume measure, this restriction does essentially nothing). If $K = 1$, we can compute the expected value of $\Omega$ using a classical stochastic gradient descent (Algorithm \ref{alg:riemannian-mean}). If $K > 1$, however, label switching may occur:  There may be a group $G$ acting on $\{1,2, \ldots, K\}$ that reindices the elements of the product $\conf{K}{M}$ without affecting likelihood.  This invalidates the expectation computed by Algorithm \ref{alg:riemannian-mean}. 

In this case, we need to work in the quotient $\conf{K}{M}\!/G$. Two key examples for $G$ will be the symmetric group $S_K$ of permutations and the cyclic group $C_K$ of cyclic permutations. When $G = S_K$ we simply recover the $K$th unordered configuration space, typically denoted $\uconf{K}{M}$.
%the quotient $\mathcal M^K/S_K$ is the \emph{symmetric power} of $\mathcal M$ in algebraic geometry \citep{macdonald1962symmetric}, also appearing as a \emph{configuration space} in physics and algebraic topology \citep{fadell2001configspaces}.

%\seb{I'm not confident about the following. I think I'm right, but I've been wrong so much that my hunches are probably false :). Proofs of these things should go in the appendix}

%Consider $\mathcal{M}^K / S_K$ specifically; the cyclic group $C_K$ and other discrete groups acting on $\mathbb Z/K$ behave similarly. 

%In a mixture model, it is usually erroneous when two components match exactly, so we  excise these consider the set $\tilde{\mathcal{M}}^K:= \mathcal{M}^K \backslash \{\}$ excise the parameter values that are not fixed by any element of $S_K$, i.e., $(p_1, \ldots, p_K)$ such that there are no repeated elements. The interior $\mathrm{int}(\mathcal{M}^K/S_K)$ is the set of  %equivalence classes of 
%vectors
%$(p_1, \ldots, p_K)$ with $p_i \neq p_j$ for any $i \neq j$, after identifying vectors that are equal up to an element of $G$. 
%where $(p_1, \ldots, p_K)$ and $(q_1, \ldots, q_K)$ are equivalent if there exists a permutation $\sigma\in S_K$ such that $p_i = q_{\sigma(i)}$ for all $i$. 
%The followi$\uthcal{M}^K/S_K)$:
%\begin{enumerate}[noitemsep]
%    \item 
$\uconf{K}{M}$ is a Riemannian manifold with structure inherited from the product metric on $\conf{K}{M}$ and has the property:
%    \item 
    \begin{equation}\label{eq:quotientdist}
        d_{\uconf{K}{M}}([(p_1, \ldots, p_K)], [(q_1, \ldots, q_K)]) = \min_{\sigma\in S_K} d_{\mathcal{M}^K}((p_1, \ldots, p_K), (q_{\sigma(1)}, \ldots, q_{\sigma(K)})).
    \end{equation}
    %\item 
    The analogous fact holds for $\mathrm{Conf}_K(\mathcal{M})/G$ for other finite $G$ via standard arguments (see e.g. \citet{kobayashi1995isometries}). Thus, we may step in the gradient direction on the quotient by solving a suitable optimal transport matching problem.
    
Since $G$ is finite, the map $\sigma$ minimizing \eqref{eq:quotientdist} is computable algorithmically.  When $G=C_K$, we simply enumerate all $K$ cyclic permutations of $(q_1,\ldots,q_K)$ and choose the one closest to $\mathbf p$.  When $G=S_K$, we can recover $\sigma$ by solving a linear assignment problem with cost $\overline c_{ij}=d(p_i,q_j)^2.$
%\end{enumerate}

% \seb{For reference. Proof sketches:
% \begin{enumerate}
%     \item $S_K$ acts freely on the set of $(p_1, \ldots, p_K)$ with $p_i \neq p_j$ for any $i \neq j$. The action is smooth and trivially proper since $S_K$ is finite. \citealt[Theorem 21.10]{lee2003introduction} then tells us that the quotient space is a topological manifold.
%     \item Positivity and symmetry are obvious. If $d_{\mathrm{Sym}}([(p_1, \ldots, p_K)], [(q_1, \ldots, q_K)]) = 0$, then there exists some permutation $\sigma$ such that $d(p_i, q_{\sigma(i)}) = 0, \forall i$ on $\mathcal{M}$. Since $d$ is a metric, this implies $p_i = q_{\sigma(i)}$, and hence $[(p_1, \ldots, p_K)] = [(q_1, \ldots, q_K)]$. Triangle inequality \emph{should} follow from gluing in transport + triangle inequality on $\mathcal{M}^K$.
%     \item $S_K$ acts by isometries; shouldn't be too hard to check. Let's say I have a Riemmanian metric $g_p(,)$ in $T_p M^K$, and $\pi : M^K \to \mathcal{M}^K/S_K$ is the quotient map. I can define a Riemmanian metric $g_{\tilde{p}}(,)$ by
%     \begin{align*}
%         \tilde{g}_{\tilde{p}}(u, v) = g_p((d\pi_p)^{-1}(u), (d\pi_p)^{-1}(v))
%     \end{align*}
%     where I can choose $p$ as any point in the fiber $\pi^{-1}(\tilde{p})$.
% \end{enumerate}

% One more thing: The restriction to the interior shouldn't be too bad. I think the boundary has codimension at least 1, so if $\Omega$ is e.g. absolutely continuous w.r.t. volume measure on $\mathcal{M}^K$, the measure of the boundary should be $0$.
% }

\begin{wrapfigure}[15]{R}[10 pt]{0.5\textwidth}
\vspace{-.3in}
\begin{minipage}{\linewidth}
\begin{algorithm}[H]
    \caption{Barycenter for Gaussian Mixtures}\label{alg:barycentersgaussian}
    \begin{algorithmic}[1]
    
    \REQUIRE{Distribution $\Omega$}
    \ENSURE {Barycenter $p= (\mu^*_1,\Sigma^*_1) \ldots, (\mu^*_K,\Sigma^*_K)$}
    \STATE{Initialize the barycenter $ p \sim \Omega$}.
    \FOR{$t=1, \ldots$}
        \STATE{Draw $((\mu_1,\Sigma_1) \ldots, (\mu_K,\Sigma_K)) \sim \Omega$}
        \STATE{Compute $\sigma$ in \eqref{eq:quotientdist}}
        \FOR{$i=1, \ldots, K$}
            \STATE{$\mu^*_i= \mu^*_i - \eta (\mu^*_i-\mu_{\sigma(i)}) $}
            \STATE{$L^*_i=L^*_i- \eta (I-T^{\Sigma^*_i \Sigma_{\sigma*(i)}})L^*_i  $}
        \ENDFOR
    \ENDFOR
    \end{algorithmic}
\end{algorithm}
\end{minipage}
\end{wrapfigure}

These properties suggest an adjustment of Algorithm \ref{alg:riemannian-mean} to account for $G$. Given a barycenter estimate $\mathbf{p} = (p_1, \ldots, p_K)$ and a draw $\mathbf{q} = (q_1, \ldots, q_K)\sim\Omega$: (1) align $\mathbf{p}$ and $\mathbf{q}$ by minimizing the right-hand side of~\eqref{eq:quotientdist}; (2) compute component-wise Riemannian gradients from $p_i$ to $q_{\sigma(i)}$; and (3) step $\mathbf{p}$ toward $\mathbf{q}$ using the exponential map. 

Algorithm \ref{alg:barycenters} summarizes our approach. It can be understood as stochastic gradient descent for $z$ in~\eqref{eq:frechetquotient}, working in  space $\conf{K}{M}$ rather than the quotient $\conf{K}{M}/G$.  Theorem~\ref{thm:quotientbarys}, however, gives an alternative interpretation.  Construct a $|G|$-point empirical distribution $\mu=\frac{1}{|G|}\sum_{\sigma\in G} \delta_{\sigma\cdot \mathbf p}$ from the iterate $\mathbf p$.  After drawing $\mathbf q\sim\Omega$, we do the same to obtain $\nu\in P_2(\conf{K}{M})$.  Then, our update can be understood as a stochastic Wasserstein gradient descent step of $\mu$ toward $\nu$ for problem~\eqref{eq:generalbarycenterobjective}.  While this equivalent formulation would require $O(|G|)$ rather than $O(1)$ memory, it imparts the theoretical perspective in \S\ref{sec:theory}, in particular a connection to the (convex) problem of Wasserstein barycenter computation.

In the supplementary, we prove the following theorem: 
\begin{thm}[Ordering Recovery]
If $\mathcal{M} = \R$, with the standard metric, then: 
    \[ \begin{aligned}\uconf{K}{M} &\cong \{(u_1, \ldots, u_K) \in \conf{K}{\R}
    \mid u_1 < \ldots < u_K \} \subset \R^K. \end{aligned} \]
Additionally, the single-orbit barycenter of Theorem \ref{thm:quotientbarys} is unique and our algorithm provably converges. 
\end{thm}
This setting occurs when one's mixture model consists of evenly weighted components with only a single mean parameter for each in $\R$. The result relates our method to the classical approach of ordering these means for correspondence and shows that it is well-justified. The convergence of our algorithm leverages the convexity of $\uconf{K}{M}$. The supplementary contains additional discussion (\S \ref{sec:positivecurvature}) about such ``mean-only'' models in $\R^d$ for $d > 1$. They lack the niceness of the $d = 1$ case, due to positive curvature. This curvature is problematic for convergence arguments (as it leads to potential non-uniqueness of barycenters), but we empirically find that our algorithm converges to reasonable results.

%We show in the appendix that this method recovers the simple ordering of means when considering an evenly-weighted model in $\R$. Additionally, we conjecture that this approach converges to a unique minimizer when $\mathcal{M}$ has non-positive sectional curvature, based on uniqueness of geodesics in non-positively curved spaces and local existence of geodesics in the quotient \citep{sturm2003probability}. 

\textbf{Mixtures of Gaussians.}
One particularly useful example %where we can compute perform Riemannian gradient descent 
involves estimating the parameters of a Gaussian mixture over $\R^d$. For simplicity, assume that all the mixture weights are equal. The manifold $\mathcal{M}$ is the set of all $(\mu, \Sigma)$ pairs: $\mathcal M\cong\R^d \times \mathcal{P}^d$ with $\mathcal{P}^d$ the set of positive definite symmetric matrices. This space can be endowed with the $W_2$ metric:
\begin{equation}
\label{eq:w2gaussians}
 d((\mu_1, \Sigma_1), (\mu_2, \Sigma_2))^2 = W_2^2 (\mathcal{N}(\mu_1,\Sigma_1), \mathcal{N}(\mu_2,\Sigma_2)) = \|\mu_1-\mu_2\|_2^2 + \mathfrak{B}^2(\Sigma_1,\Sigma_2),
\end{equation}
where $\mathfrak{B}^2$ is the squared Bures metric 
$\mathfrak{B}^2(\Sigma_1,\Sigma_2)= \mathrm{Tr}[\Sigma_1+\Sigma_2-2(\Sigma_1^{\frac{1}{2}}\Sigma_2\Sigma_1^{\frac{1}{2}})^{\frac{1}{2}} ].$ 

As the mean components inherit the structure of Euclidean space, we only need to compute Riemannian gradients and exponential maps for the Bures metric. \citet{muzellec2018} leverage the Cholesky decomposition to parameterize $\Sigma_i = L_i L_i^\intercal$. The gradient of the Bures metric then becomes:
%$$\begin{aligned}&\nabla_{L_1}\frac{1}{2}\mathfrak{B}(\Sigma_1,\Sigma_2) = (I-T^{\Sigma_1 \Sigma_2})L_1 \\
%&\quad \text{with} \quad T^{\Sigma_1 \Sigma_2}= \Sigma_1^{-\frac{1}{2}} (\Sigma_1^{\frac{1}{2}}\Sigma_2\Sigma_1^{\frac{1}{2}})^{\frac{1}{2}}\Sigma_1^{-\frac{1}{2}}. \end{aligned} $$ 
$$\nabla_{L_1}\frac{1}{2}\mathfrak{B}(\Sigma_1,\Sigma_2) = (I-T^{\Sigma_1 \Sigma_2})L_1 
\quad \text{with} \quad T^{\Sigma_1 \Sigma_2}= \Sigma_1^{-\frac{1}{2}} (\Sigma_1^{\frac{1}{2}}\Sigma_2\Sigma_1^{\frac{1}{2}})^{\frac{1}{2}}\Sigma_1^{-\frac{1}{2}}$$
The 2-Wassertein exponential map for SPD matrices  is $\exp_{\Sigma}(\xi) = (I+\mathcal{L}_\Sigma(\xi)) \Sigma (I+\mathcal{L}_\Sigma(\xi))$ where $\mathcal{L}_\Sigma(\xi)$ is the solution of this Lyapunov equation : $\mathcal{L}_\Sigma(\xi)\Sigma+\Sigma \mathcal{L}_\Sigma(\xi) = \xi$.

\begin{wrapfigure}[14]{r}{0.4 \linewidth} %{lin}
    \centering
        %\includegraphics[scale=0.3]{images/butterfly.pdf}
    %\includegraphics[scale=0.3]{images/Ellipse-1-new.pdf}
    %\includegraphics[scale=0.3]{images/Ellipse-2-new.pdf}
    %\includegraphics[scale=0.3]{images/Ellipse-3-new.pdf}
    %\includegraphics[scale=0.3]{images/Ellipse-4-new.pdf}
    %\includegraphics[scale=0.3]{images/Ellipse-5-new.pdf}
    %\raisebox{0pt}[0in\relax]{
    \vspace{-0.7in}
    \includegraphics[width=1in]{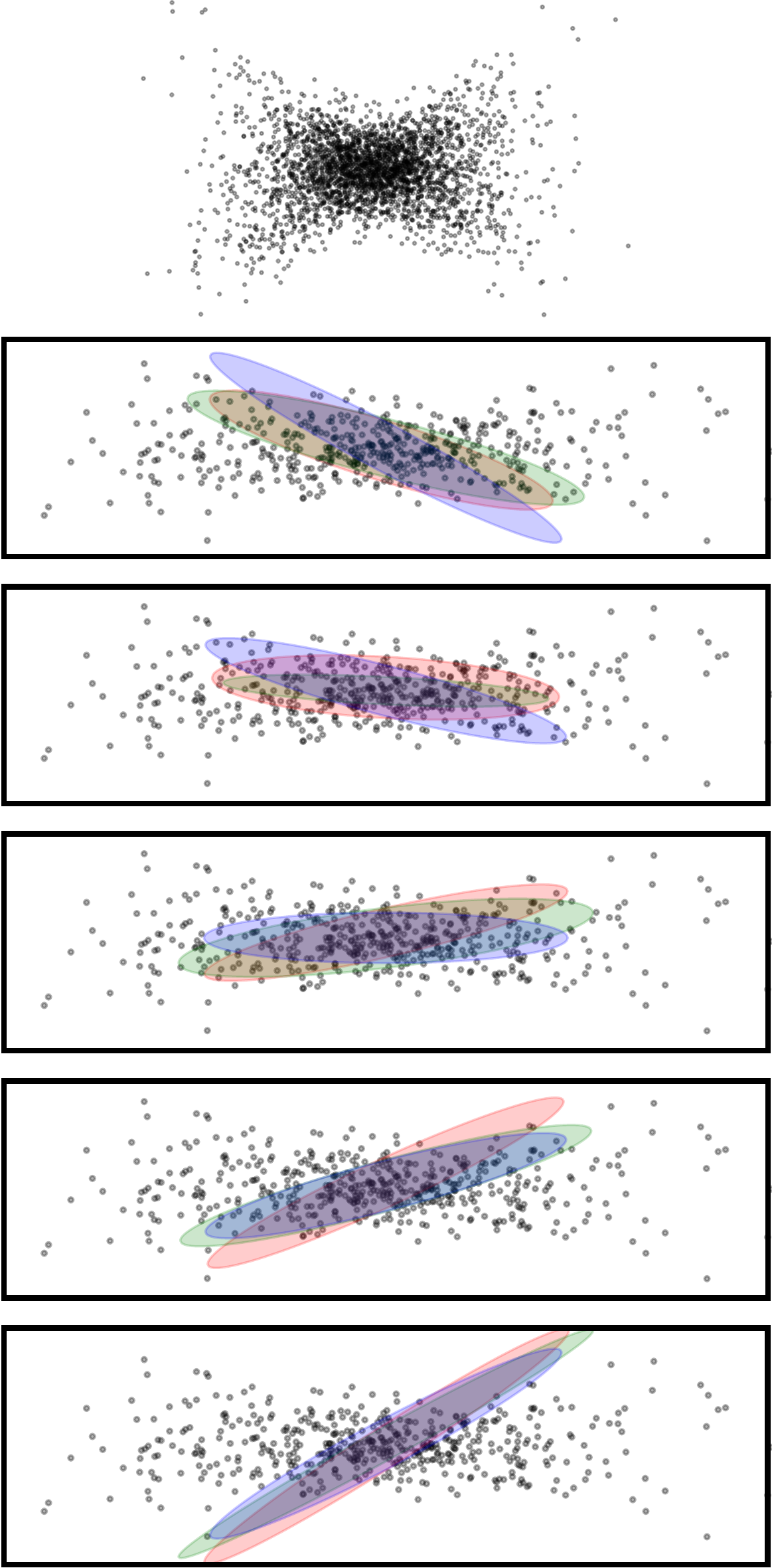}
    %}
    \caption{True covariances in {\color{bleu!51}   blue}, covariances from SGD in {\color{vert!51}  green} and pivot in {\color{rouge!51}  red }}
    \label{fig:ellipse}
\end{wrapfigure}

\section{Results}
\label{sec:results}
In \S\ref{sec:algorithms}, we gave a symmetry-invariant, simple, and efficient algorithm for computing a Wasserstein barycenter to summarize a distribution subject to label switching. To verify empirically that our algorithm can efficiently address label switching, we test on two natural examples: estimating the parameters of a Gaussian mixture model and a Bayesian instance of multi-reference alignment.

%In the previous section, we derived an efficient algorithm that takes advantage of working with Gaussian mixture and gives a way to average samples that cannot be averages directly. We therefore applied our Algorithm \ref{alg:barycenters} to synthetic and real data to show its performance. We first apply it to the of raw output  of an HMC sampler, then we proved that our method may break some baselines. Afterwords, we apply our method to the case of the cyclic group and finally we used to generated new handwritten digits.
%\begin{wrapfigure}{R}{0.5 \textwidth}
%    \centering
%    \includegraphics[scale=0.2]{images/figures/HMCoutput.pdf}
%    \includegraphics[scale=0.2]{images/figures/Barycenter-trajectory.pdf}
%    \includegraphics[scale=0.2]{images/figures/HMCoutput-pivot.pdf}
%    \includegraphics[scale=0.2]{images/figures/HMCoutput-steph.pdf}
%    \caption{Left: HMC raw output, barycenter trajectory computed via SGD, Pivot and Stephens's relabeling output}
%    \label{fig:HMC}
%\end{wrapfigure}
%\begin{wrapfigure}[20]{r}{0.45\textwidth}
%\vspace{-.2in}
 %   \centering
    %\includegraphics[scale=0.3]{images/butterfly.pdf}
    %\includegraphics[scale=0.3]{images/Ellipse-1-new.pdf}
    %\includegraphics[scale=0.3]{images/Ellipse-2-new.pdf}
    %\includegraphics[scale=0.3]{images/Ellipse-3-new.pdf}
    %\includegraphics[scale=0.3]{images/Ellipse-4-new.pdf}
    %\includegraphics[scale=0.3]{images/Ellipse-5-new.pdf}
   % \includegraphics[scale=0.3]{images/drawing-1.pdf}
  %  \caption{True Covariances, Covariances from SGD and the Pivot}
    %\label{fig:ellipse}
%\end{wrapfigure}

\textbf{Estimating components of a Gaussian mixture.}
Our first scenario is estimating the parameters of a Gaussian mixture with $K > 1$ components. We use Hamiltonian Monte Carlo (HMC) to sample from the posterior distribution of a Gaussian mixture model. Na\"ive averaging does not yield a meaningful barycenter estimate, since the samples are not guaranteed to have the same label ordering. %(see Figure \ref{fig:HMC}). 
To resolve this ambiguity, we apply our method and two baselines: the pivotal reordering method  \citep{marin2005bayesian} and Stephens' method \citep{Stephens2000}.  The Stephens and Pivot methods relabel samples. Stephens minimizes the Kullback–Leibler divergence between average classification distribution and classification distribution of each MCMC sample. Pivot aligns every sample to a prespecified sample (i.e. pivot) by solving a series of linear sum assignment problems. Pivot method requires pre-selecting a single sample for alignment — poor choice of the pivot sample leads to bad estimation quality, while making a “good” pivot choice may be highly non-trivial in practice. The default pivot choice is the MAP. Stephens method is more accurate, however it is expensive computationally and has large memory requirement.

%The two baselines have their own strengths and weaknesses: the pivot reordering method is fast, but may not perform well depending on the choice of pivot, while Stephens' method computes very accurate relabelings, but is computationally taxing. Our proposed stochastic gradient approach sits in between these two extremes: it is computationally faster than Stephens', and more accurate than pivotal reordering.

\begin{wraptable}[5]{r}{0pt}
\centering
 \raisebox{7pt}[\dimexpr\height-1.5\baselineskip\relax]{
 \begin{tabular}{@{}l@{\quad}r@{\quad}r@{\quad}r}
  \toprule
    & Pivot & Stephens & SGD \\
   \midrule
   {\bfseries Error (abs)} & 1.65 & 1.26 & 1.47  \\
   {\bfseries Time  (s)} & 1.4 &  54  &  7.5   \\
   \bottomrule
 \end{tabular}
 }
%  \vspace*{-2.5cm}
 \caption{Absolute error \& timings}
 \vspace{-0.08in}% for a mixture of Gaussians}
 \label{tab:error-butter}
 \end{wraptable}

%\begin{figure*}
%\begin{floatrow}
%\ffigbox{%
%  \centering
%  \begin{tabular}{@{}c@{}c@{}}
%    \includegraphics[scale=0.4]{images/error-without-error-HMC-samples-new.pdf} &
%    \includegraphics[scale=0.4]{images/error-without-error-HMC-time-new.pdf}
%    \end{tabular}
%}{%
%  \caption{Relative error as a function of  (a) number of samples and (b) time.}\label{fig:err-HMC}
%}\qquad
%\capbtabbox{%
% \centering\small
%\raisebox{0pt}[\dimexpr\height-1.5\baselineskip\relax]{
%\begin{tabular}{@{}l@{\quad}r@{\quad}r@{\quad}r}
%   \toprule
%    & Pivot & Stephens & SGD \\
%  \midrule
%   {\bfseries Error (abs)} & 1.65 & 1.26 & 1.47  \\
%   {\bfseries Time  (s)} & 1.4 &  54  &  7.5   \\
%   \bottomrule
%\end{tabular}
%}
%}{%
%  \caption{Absolute error \& timings}\label{tab:error-butter}
%}
%\vspace{-.2in}
%\end{floatrow}
%\end{figure*}
To illustrate why pivoting fails, consider samples drawn from
%the example in Figure \ref{fig:err-HMC}. The data 
a mixture of five Gaussians with mean $0$ and covariances $R_\theta M$ with $M= \bigl(\begin{smallmatrix} 
1& 0 \\
0 & 0.1 
\end{smallmatrix}\bigr)$ and $R_\theta$ a rotation of angle $\theta\in \lbrace -\nicefrac{\pi}{12}, -\nicefrac{\pi}{24}, 0, \nicefrac{\pi}{12}, \nicefrac{\pi}{24}\rbrace$ (Figure \ref{fig:ellipse}). The resulting pivot is uninformative for certain components. The underlying issue is that the pivot is chosen to maximize the posterior distribution. If this sample lies on the boundary of $\conf{K}{M} / S_K$, the pivot cannot be effectively used to realign samples. Quantitative results for this test case are in Table \ref{tab:error-butter}.

%The main drawback of the pivot method was the choice of this privileged sample. The pivot method fails when the choice of this sample does not cope with the symmetries and overlapping that may exists in the space of parameters. We give here two examples. 

%The first one was the case of a distribution made of five components all centered in (0,0) with covariances high in one direction and small in the other and images from one to another of a rotation of small angle (Figure \ref{fig:ellipse}). The HMC was performed with four chains of 400 iterations (200 warm-up and 200 sampling) and with 100 points of data points. Pivot method had a bigger error in this case (Table \ref{tab:error-butter}) because the privileged sample is not aligned with the true covariances. As in the relabeling, the pivot method aligns all the sample to this particular sample, and some misalignment may occur, as the different covariances cluster overlap. Therefore, the samples average after pivot relabeling gives us bigger error than the SGD output. 

% \begin{wrapfigure}{r}{0pt}
% \vspace{-.3in}
%     \centering
%     \includegraphics[scale=0.25]{images/error-without-error-HMC-samples-new.pdf}
%     \includegraphics[scale=0.25]{images/error-without-error-HMC-time-new.pdf}
%     \caption{Relative error as a function of  (a) number of samples and (b) time.}
%     \label{fig:err-HMC}
% \end{wrapfigure}

\begin{figure}[H]
\vspace{-.2in}
     \centering
     \includegraphics[scale=0.3]{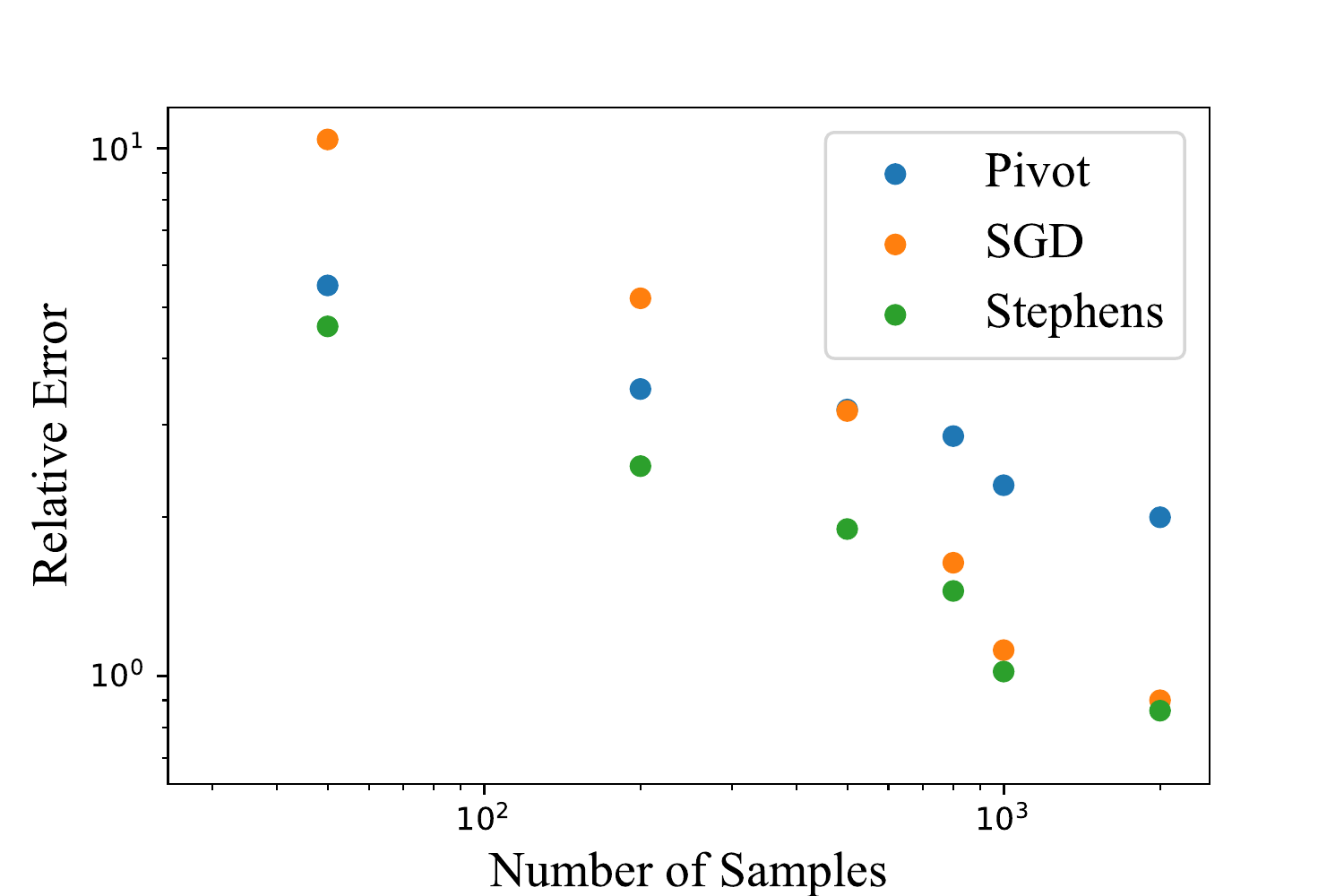}
     \includegraphics[scale=0.3]{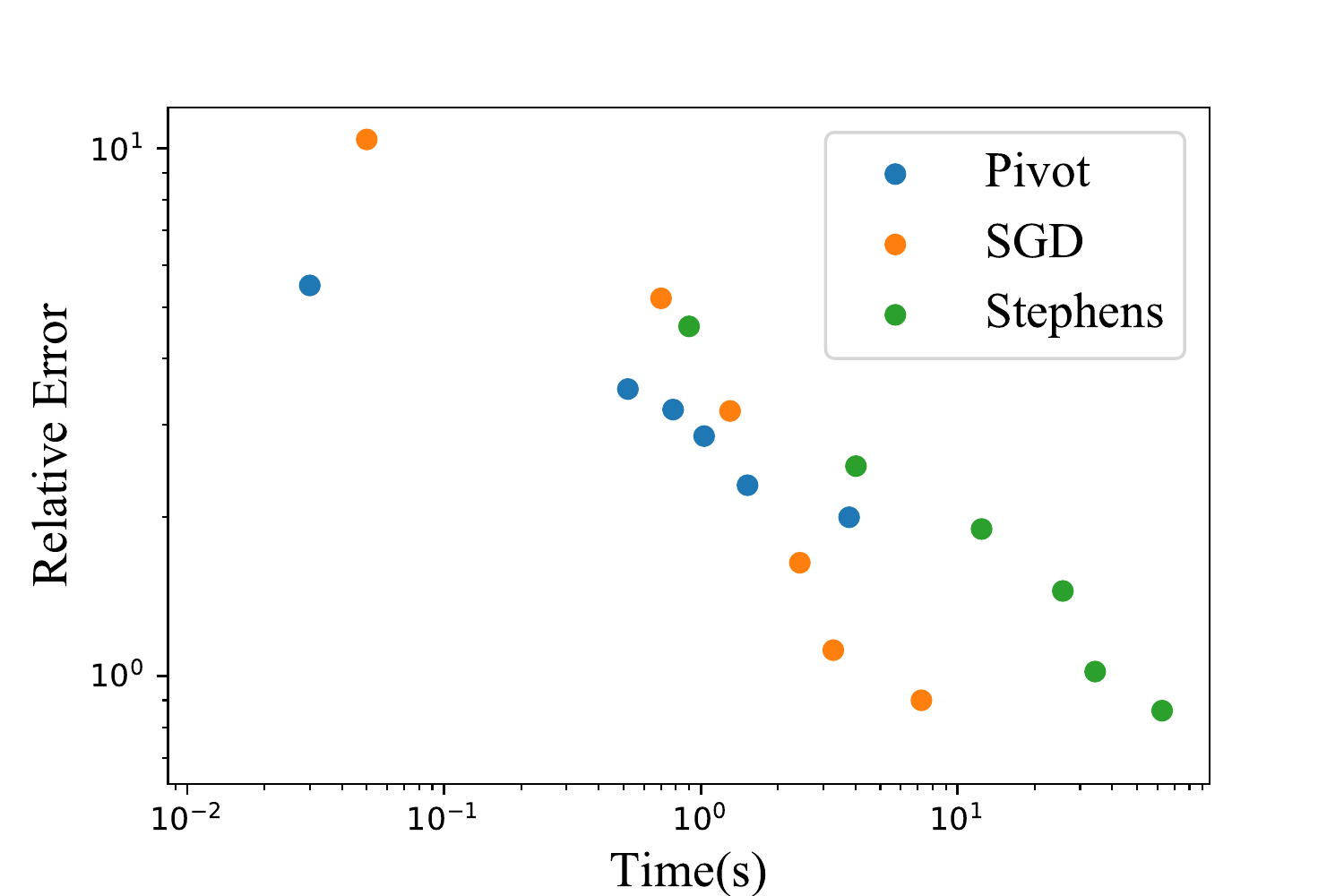}
     \vspace{.1in}
     \caption{Relative error as a function of  (a) number of samples and (b) time.}
     \vspace{-.15in}
     \label{fig:err-HMC}
\end{figure}

To get a better handle of the performance/accuracy trade-off for the three methods, we run an additional experiment. We draw samples from a mixture of five Gaussians over $\R^5$ with means $0.5 e_i$, where $e_i\in\R^5$ is the $i$-th standard basis vector with $i\in\{1,\ldots,5\}$, and covariances $0.4 I_{5\times 5}$. %that have coordinates 0 in all the directions except one in which coordinate equals $0.5$ (every component corresponds to a different direction) and 0.4 times identity covariances. 
We implement HMC sampler using \texttt{Stan} \citep{carpenter2017stan}, with four chains discarding 500 burn-in samples and keeping $500$ per chain. Then we compare the three methods, increasing the number of samples to which they have access. We measure relative error as a function of  wall clock time and number of samples (Figure \ref{fig:err-HMC}). The resulting plots align with our intuition: pivoting obtains a suboptimal solution quickly, but if a more accurate solution is desired, it is better to run our SGD algorithm.

\vspace{-.1in}
%\seb{The plot needs to prove this :)}

%The second experiment to compare SGD performances to baselines was a five dimensional five components GMM. The means of each component have coordinates 0.3 in one direction and zero in the others and the covariances were $I_2$, so that the distribution overlaps one each other. The number of points that were taken for HMC was 120. As shown in Figure \ref{fig:err-HMC} left , the longest part to estimate the parameters of our GMM is the HMC sampling. As expected, the pivot method is very fast, the Stephen's pretty slow and the SGD is in between but closer to the pivot performances. Also, the performances of SGD is always better than the pivot method and very much better when a small numbers of points are sampled from the posterior distribution due to the reason exposed in the previous experiment. The SGD performs worse than the Stephen's method in terms of accuracy but outperforms in terms of time.  SGd appears to be a good trade-off between accuracy and time, better than the Pivot and the Stephen's methods. 

\begin{figure}[h]
    \centering
    \vspace{-.1in}
    \includegraphics[scale=0.3]{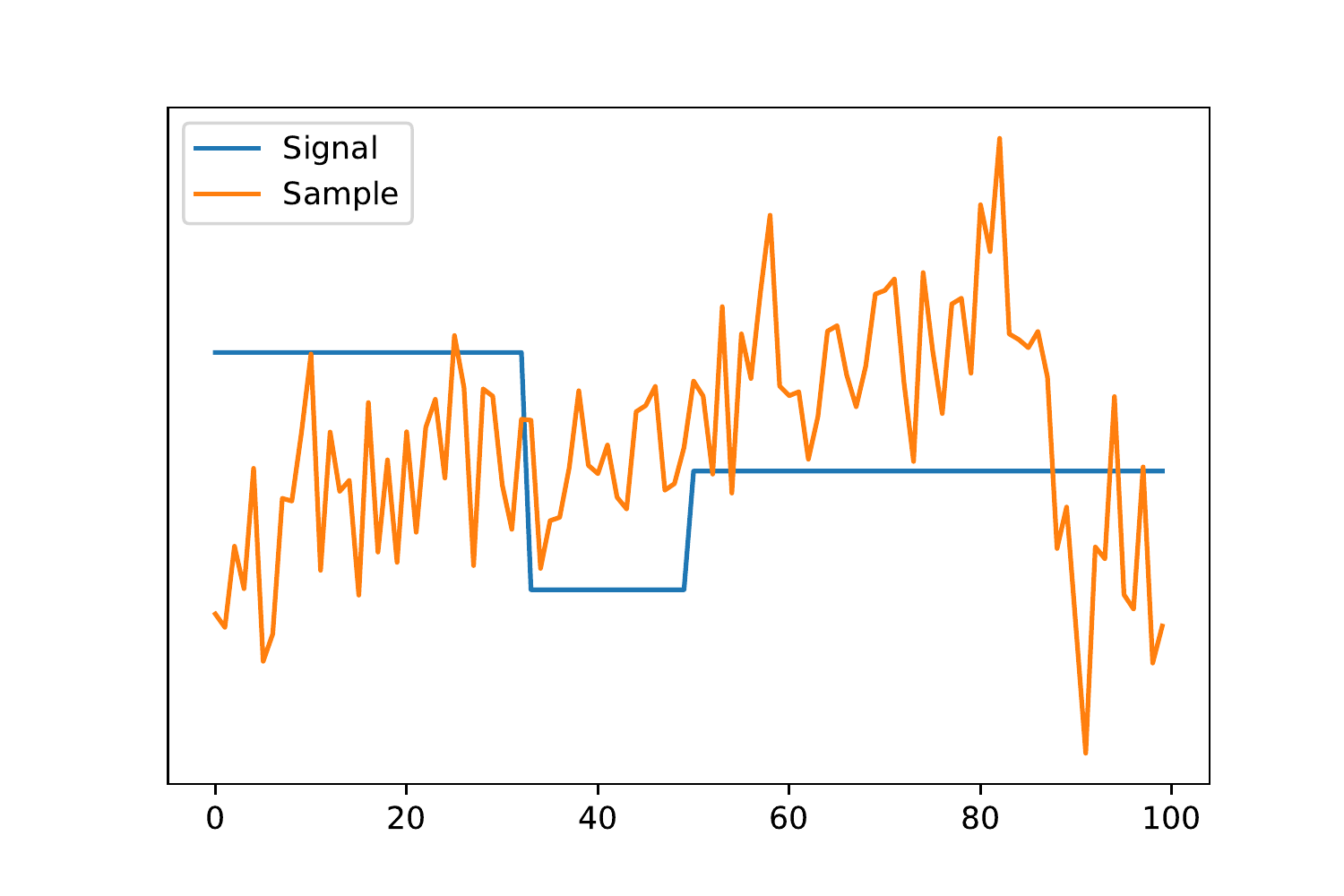}
    \includegraphics[scale=0.3]{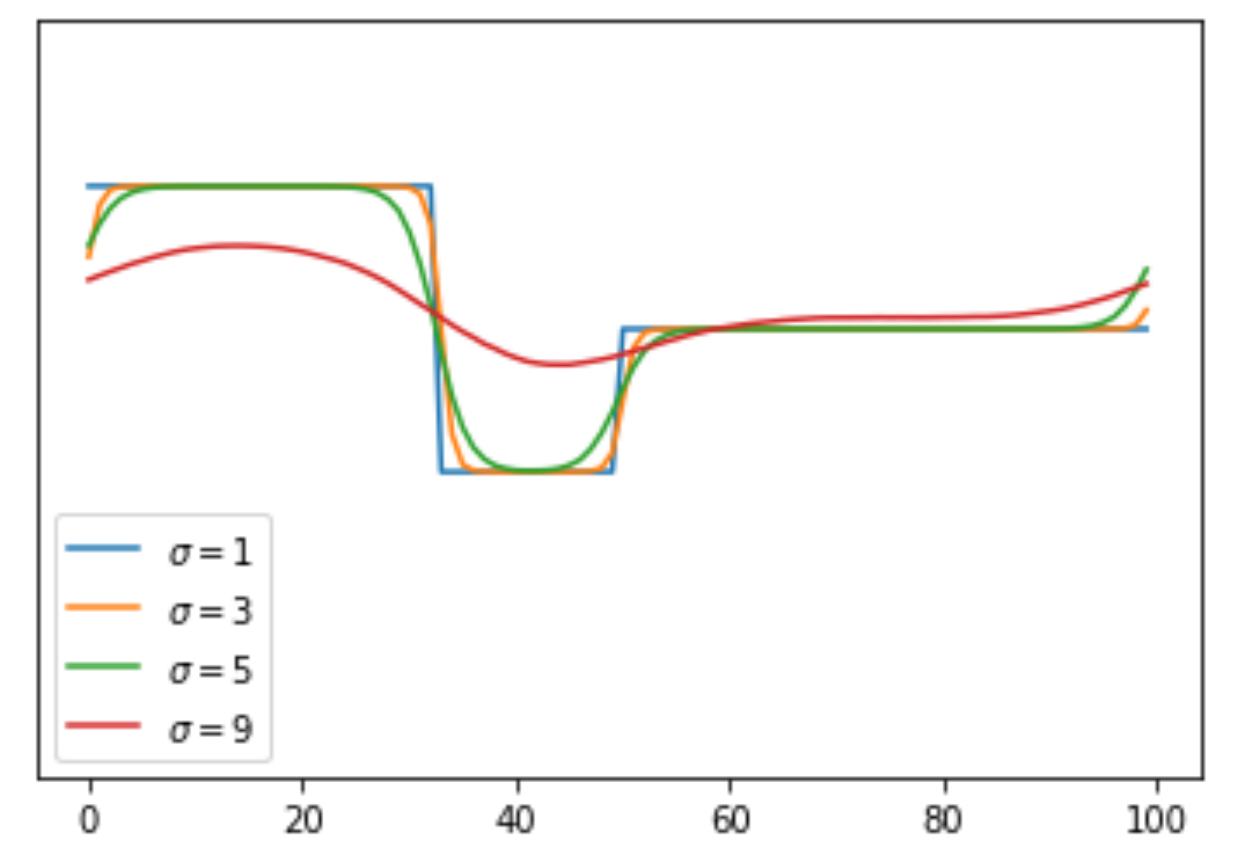}
    \includegraphics[scale=0.3]{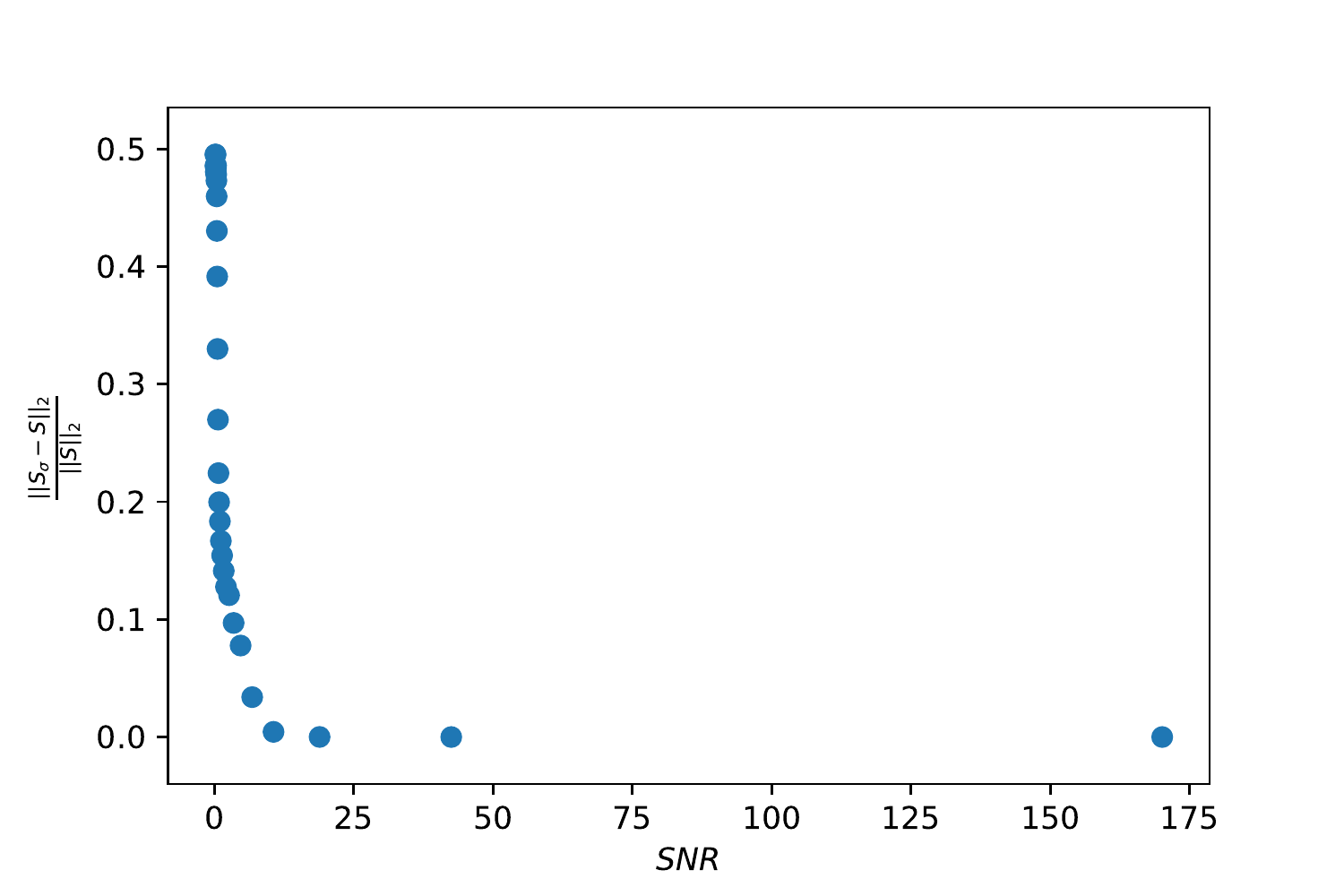}
    \caption{Reconstruction of a signal from shifted and noisy observations. (a) The true signal is plotted in blue against a random shifted and noisy draw from the MCMC chain. (b) Reconstructed signals at varying values of noise. (c) Relative error as a function of SNR.}
    \vspace{-.1in}
    \label{fig:cyclic-group}
\end{figure}

\textbf{Multi-reference alignment.}
A different problem to which we can apply our methods is \emph{multi-reference alignment} \citep{zwart2003fast,bandeira2014multireference}. We wish to reconstruct a template signal $x \in \R^K$ given noisy and cyclically shifted samples $y \sim g\cdot x + \mathcal{N}(0, \sigma^2 I),$
where $g \in C_K$ acts by cyclic permutation. These observations correspond to a mixture model with $K$ components $\mathcal{N}(g\cdot x, \sigma^2 I)$ for $g\in C_K$ \citep{perryweed17}. We simulated draws from this distribution using Markov Chain Monte Carlo (MCMC), where each draw applies a random cyclic permutation and adds Gaussian noise (Figure \ref{fig:cyclic-group}a). The sampler we used was a Gibbs Sampler \citep{Casella1992}.
To reconstruct the signal, we first compute a barycenter using Algorithm \ref{alg:barycenters}, giving a reference point to which we can align the noisy signals; we then average the aligned samples. Reconstructed signals for different $\sigma$'s are in Figure \ref{fig:cyclic-group}b. To evaluate quantitatively, we compute the relative error of the reconstruction as a function of signal-to-noise ratio $\mathrm{SNR} = \nicefrac{\|x\|^2}{K \sigma^2}$ (Figure \ref{fig:cyclic-group}c).

\vspace{-.1in}

\section{Discussion and Conclusion}
\label{sec:conclusions}

The issue underlying label switching is the existence of a group acting on the space of parameters. This group-theoretic abstraction allows us to relate a widely-recognized problem in Bayesian inference to Wasserstein barycenters from optimal transport. Beyond theoretical interest, this connection suggests a well-posed and easily-solved optimization method for alleviating label switching in practice.

The new structure we have revealed in the label switching problem opens several avenues for further inquiry. Most importantly, \eqref{eq:frechetquotient} yields a simple algorithm, but this algorithm is only well-understood when the Fr\'echet mean is unique. This leads to two questions: When can we prove uniqueness of the mean? More generally, are there efficient algorithms for computing barycenters in $P_2(X)^G$?

%The first question is of theoretical interest, since the group in Bayesian inference is often a symmetry group $S_K$, and thus we seek only to characterize spaces $X$ for which $X/G$ is well behaved. \justin{previous sentence is a run-on that I had some trouble fixing; also wasn't sure how the two halves are related} As concrete directions, we can ask whether assuming convexity and non-negative curvature of $X$ is sufficient to yield a uniqueness result.

Finding faster algorithms for computing barycenters under the constraints of Lemma \ref{lem:barys} provides an unexplored and highly-structured instance of the barycenter problem. Current approaches, such as those by \citet{cuturi_fast_2014} and \citet{DBLP:conf/icml/ClaiciCS18} are too slow and not tailored to the demands of our application, since each measure is supported on $K!$ points and the barycenter may not share support with the input measures.  Moreover, after incorporating an HMC sampler or similar piece of machinery, our task likely requires taking the barycenter of an infinitely large set of distributions. The key to this problem is to exploit the symmetry of the support of the input measures and the barycenter.

%We have made an important step in resolving the label switching problem by joining together Bayesian inference and optimal transport. Our results show that combining insight from these two fields can lead to practical and principled algorithms.%<---- sort of a repeat of the first paragraph

%Why does the label switching problem occur? In Bayesian mixture models, exchangeable priors lead to indistinguishable regions in the posterior likelihood, but the underlying problem 

\paragraph*{Acknowledgements.}
J.\ Solomon acknowledges the generous support of Army Research Office grant W911NF1710068, Air Force Office of Scientific Research award FA9550-19-1-031, of National Science Foundation grant IIS-1838071, from an Amazon Research Award, from the MIT-IBM Watson AI Laboratory, from the Toyota-CSAIL Joint Research Center, from the QCRI--CSAIL Computer Science Research Program, and from a gift from Adobe Systems. Any opinions, findings, and conclusions or recommendations expressed in this material are those of the authors and do not necessarily reflect the views of these organizations.

\bibliography{transport_refs}
\bibliographystyle{icml2019}

\clearpage
\appendix

\section{Optimal Transport Theory}
\subsection{Proof of Theorem 1}
\label{sec:prokhorov}
We first recall the definition of sequential compactness and Prokhorov's theorem, which relates it to tightness of measures:
\begin{defn}[Sequential compactness]
    A space $X$ is called \emph{sequentially compact} if every sequence of points $x_n$ has a convergent subsequence converging to a point in $X$.
\end{defn}

\begin{thm}[Prokhorov's theorem]
    A collection $C \subset P_2(X)$ of probability measures is tight if and only if $C$ is sequentially compact in $P_2(X)$, equipped with the topology of weak convergence.
\end{thm}

Now, note that the barycenter objective is bounded below by 0 and is finite, so we may pick out a minimizing sequence $\mu_n$ of $B(\mu)$. Prokhorov's theorem allows us to extract a subsequence $\mu_{n_k}$ that converges to a minimizer $\mu \in P_2(X)$ and the theorem is proved.

\subsection{Tightness from Uniform Second Moment Bound}
\label{sec:tightness}

We argue here for a sufficient condition for tightness claimed in the text:
\begin{lem}
%     The problem
% \begin{equation}
%     \label{eq:barys}
%     \inf_{\mu \in P_2(X)} \mathbb{E}_{\nu\sim \Omega}\left[W_2^2(\mu, \nu)\right].
% \end{equation}
If a collection of measures $\mathcal{C} \subset P_2(X)$ has a uniform second moment bound (about any reference point $x_0 \in X$), i.e.,
\[ \int_X d^2(x_0,x) \, d\nu(x) < M \]
for some $M > 0$ and all $\nu \in \mathcal{C}$, then $\mathcal{C}$ is tight.
\end{lem}
\begin{proof}
    For any $\nu \in \mathcal{C}$ we have the following inequalities:
    \begin{align*}
        \nu\{x\ |\ d(x,x_0) > R\} = \int_{d(x,x_0) > R} \mathrm{d}\nu \leq \frac{1}{R^2} \int_{d(x,x_0) > R} d(x,x_0)^2\mathrm{d}\nu(x) \leq \frac{M}{R^2}.
    \end{align*}
    The last term converges to $0$ as $R\to \infty$, and the set $\{x\ |\ d(x,x_0) \leq R\}$ is compact, so tightness follows.
\end{proof}

\subsection{Mean-only Mixture Models} \label{sec:meanonly}

Here we note some facts about mixture models, where the $K$ components are evenly weighted and identical with only one parameter each in $\R^d$. An example would be the simple case of a Gaussian mixture model with fixed equal covariance across each component, and a remaining unspecified mean parameter $p_i \in \R^d$.

\begin{figure}
\centering
\def\svgwidth{.3\columnwidth}
\graphicspath{{./figures/}}
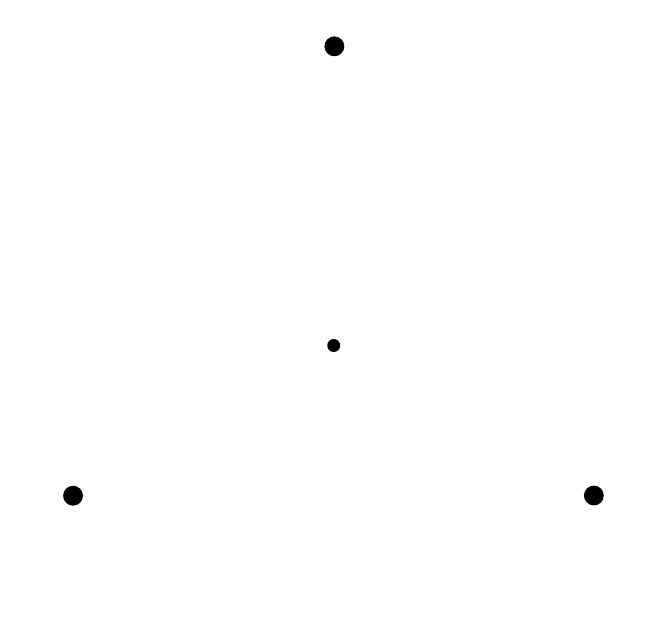
\caption{A schematic illustrating the nontrivial part of the action of $S_3$ on $\R^3$. It acts on $F_3^\bot$ and the embedded $2$-simplex shown via reflection over the dashed lines. One can see that reflection over these lines correspond to swapping of pairs of means, generating $S_3$ as a group. } \label{fig:R3S3}
\vspace{-.15in}
\end{figure}

In this instance, we are taking the quotient of $(\R^d)^K$ by an action of $S_K$ which simply permutes the $K$ factors of the product. Let us begin by investigating the case where $d = 1$. In this instance, we note that the sum of the scalar means $\sum_i p_i$ remains fixed under the action of the group. In fact, the action of the group splits into a trivial action on the 1-dimensional fixed subspace $F_K := \{(p_1, \ldots, p_k) \mid p_i \, \mathrm{all} \, \mathrm{equal} \}$, and an action on $F_K^\bot$ which permutes the vertices of an embedded regular $(K-1)$-simplex about the origin. Namely, one may take the simplex in $F_K^\bot$ with vertices that consist of the point $(K-1, -1, -1, \ldots, -1)$ and its orbit. Figure \ref{fig:R3S3} illustrates the concrete example of three means: $\R^3 / S_3$. It shows $F_3^\bot$, an embedded $2$-simplex, and the action of $S_3$ on this space and simplex. Section \ref{sec:thm3proof} proves that the quotient space $\R^K / S_K$ is a convex, easily described set, and discusses the consequences for label switching.

The splitting mentioned above is the decomposition into irreducible components. For $d>1$, the action of $S_K$ is diagonal and acts on the $d$ components of the means $p_i$ in parallel. It preserves the scalar sum of these components over each dimension and we obtain the following splitting for the general case:
\begin{equation} (\R^d)^K = \bigoplus_{j=1}^d \left(F_K \oplus F_K^\bot \right) \cong \R^d \oplus \left( \R^{K-1} \right)^d.\label{eq:splitting}\end{equation}
The action on the first $\R^d$ component is trivial, while the second component has the diagonal action permuting the vertices of an embedded regular $(K-1)$-simplex for each $\R^{K-1}$. The simple example of two means in $\R^2$ ($d = K = 2$) is discussed and illustrated in the next section (\ref{sec:uniquenesscounter}), and also serves to provide a counterexample to barycenter uniqueness. For $d>1$, the quotient $(\R^d)^K / S_K$ lacks the simple convexity of the $d=1$ case, as described in Section \ref{sec:positivecurvature}.

\subsection{Counterexample to uniqueness} \label{sec:uniquenesscounter}
\begin{wrapfigure}[6]{r}{.12\linewidth}\centering
% \raisebox{-0.5\height}{\includegraphics[scale=0.5]{imgs/RWMDFails2a.png}}\hspace{1in}
% \raisebox{-0.5\height}{\includegraphics[scale=0.5]{imgs/RWMDFails2b.png}}
\vspace{-.1in}
\centering
\def\svgwidth{\linewidth}
\graphicspath{{./figures/}}\scriptsize
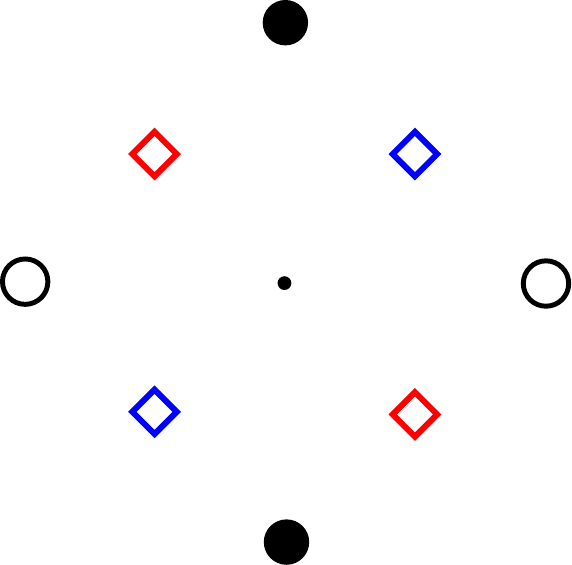
\label{fig:barycenterNonunique}
\end{wrapfigure}
Take $d = K = 2$ from the scenario above, which might correspond to our mixture model consisting of two Gaussians in $\R^2$ with equal weights and fixed variance. Only the means $(x,y; z,w) \in (\R^2)^2$ are taken as parameters, and the action of $S_2$ swaps the means: $(x,y; z,w) \mapsto (z,w; x,y)$. This action splits into a trivial action on $\mathrm{Span}\{(1,0;1,0),(0,1;0,1)\}$ and an antipodal action ($v \mapsto -v$) on $\mathrm{Span}\{(1,0;-1,0),(0,1;0,-1)\}$, where these are the first and second components in Eq. \eqref{eq:splitting}. Recall that the 1-simplex is just an interval and the action of $S_2$ merely flips the endpoints, so the antipodal action arises as the diagonal action of this flip.

The inset figure illustrates a simple schematic counterexample in the second span. The two distributions to be averaged are evenly supported on the black and white dots, invariant under reflection through the center origin $O$. Two candidate barycenters are those evenly supported on the red and blue diamonds, and in fact, any convex combination of these two are a barycenter. This corresponds to averaging a mixture with means at $(1,0)$ and $(-1,0)$ and another with means at $(0,1)$ and $(0,-1)$. Two sensible averages are a pair of means at $(0.5,0.5)$ and $(-0.5,-0.5)$, or a pair of means at $(0.5, -0.5)$ and $(-0.5,0.5)$.

Note that the previous example requires a high degree of symmetry for the input distributions, and uniqueness is recovered if either of the distributions are absolutely continuous. Section \ref{sec:positivecurvature} further characterizes the geometry of the quotient space for $d=K=2$, and how it leads to non-unique barycenters.

\section{Optimal Transport with Group Invariances}

\subsection{Proof of Lemma 4}
\label{sec:lemma4}

Consider an arbitrary point $z_0 \in X/G$, and we will show that a minimizer of $z \to \mathbb{E}_{\delta_x\sim \Omega_*} \left[d(x, z)^2\right]$ lies in a closed ball about $z_0$. As the function is continuous and this is a compact set, existence of a minimizer results.

By the triangle inequality, we have $d(x,z) \geq d(x,z_0) - d(z,z_0)$. Thus, we have:
\begin{align*}
    \mathbb{E}_{\delta_x\sim \Omega_*}\left[d(x,z)^2\right] &= \int_{X/G} d(x, z)^2 \,\mathrm{d}\Omega_*(\delta_x)\\
    &\geq \int_{X/G} \left(d(x, z_0) - d(z,z_0)\right)^2 \,\mathrm{d}\Omega_*(\delta_x)\\
    &= \left( \int_{X/G} d(x, z_0)^2 \,\mathrm{d}\Omega_*(\delta_x)\right) + d(z,z_0)^2 - 2d(z,z_0) \int_{X/G} d(x,z_0) \,\mathrm{d}\Omega_*(\delta_x).\\
\end{align*}
The last two terms are quadratic in $d(z,z_0)$. Given an arbitrary positive constant $M > 0$, some simple algebra shows that:
\[ d(z,z_0) > \frac{c + \sqrt{c^2 + 4M}}{2} \implies d(z,z_0)^2 - c d(z,z_0) > M \]
where $c = 2\int_{X/G} d(x,z_0) \,\mathrm{d}\Omega_*(\delta_x)$. The finiteness of this integral follows from the fact that $\Omega_*$ has finite second moment, implying finite first moment. Thus, if we set $M$ to a realized value of $\mathbb{E}_{\delta_x\sim \Omega_*} \left[d(x, z)^2\right]$, we see that a minimizer lies in the ball of radius $\frac{c + \sqrt{c^2 + 4M}}{2}$ about $z_0$. Taking $z$ outside this ball implies:
\begin{align*}
    \mathbb{E}_{\delta_x\sim \Omega_*}\left[d(x,z)^2\right] &\geq \left( \int_{X/G} d(x, z_0)^2 \,\mathrm{d}\Omega_*(\delta_x)\right) + d(z,z_0)^2 - 2d(z,z_0) \int_{X/G} d(x,z_0) \,\mathrm{d}\Omega_*(\delta_x).\\
    &\geq d(z,z_0)^2 - 2d(z,z_0) \int_{X/G} d(x,z_0) \,\mathrm{d}\Omega_*(\delta_x) > M.
\end{align*}

\subsection{Proof of Theorem 3} \label{sec:thm3proof}

We recall the minimization problem in (5) of the paper for a sample  $\mathbf{q} = (q_1, \ldots, q_K)$ and a current barycenter estimate $\mathbf{p} = (p_1, \ldots, p_K)$  (with a squared distance objective for simplicity of expression):
    \begin{equation}\label{eq:quotientdist}
        \min_{\sigma\in S_K} d^2_{\mathcal{\R}^K}((p_1, \ldots, p_K), (q_{\sigma(1)}, \ldots, q_{\sigma(K)})) = \min_{\sigma \in S_K} \sum_{i=1}^K \|p_i - q_{\sigma(i)} \|^2.
    \end{equation}
Here, we invoke the monotonicity of transport in 1D (see e.g. \cite{santambrogio_optimal_2015}, Chapter 2) to see that we should simply order $\mathbf{q}$ in the same way that $\mathbf{p}$ is. That is to say: assuming $p_1 < p_2 < \ldots < p_K$ (WLOG), then the optimal $\sigma$ is such that $q_{\sigma(1)} < q_{\sigma(2)} < \ldots < q_{\sigma(K)}$.

The above argument also shows that we have a very concrete realization:
\[\uconf{K}{\R} \cong \{(u_1, \ldots, u_K) \in \conf{K}{\R} \mid u_1 < \ldots < u_K \} \subset \R^K .\]

As this is an open convex set, we have uniqueness of the single-point barycenter of Theorem 2 from the paper under mild conditions on the posterior. Namely, consider that $\Omega_* \in P_2(P_2(X))$ descends to a measure $\Omega_\downarrow \in P_2(X)$, and we will need to assume that $\Omega_\downarrow$ is absolutely continuous (as you might expect). With this, \citet{kim_wasserstein_2017} give us the desired result.

Furthermore, we have guaranteed convergence of stochastic gradient descent (our algorithm) in this setting, as $\mathbb{E}[W^2_2(\cdot,\nu)]$ is 1-strongly convex and the domain is convex.
The next section shows us that we may not leverage such simple structure for $d > 1$.

\subsection{Positive Curvature of Mean-Only Models} \label{sec:positivecurvature}

Section \ref{sec:uniquenesscounter} shows us that in the case of $d = K = 2$:
\[ \uconf{2}{\R^2} \cong \R^2 \times C^* \qquad \mathrm{where} \qquad C^* = (\R^2 \backslash \{(0,0)\}) / \{ v \sim -v \}. \]
$C^*$ is isometric to an infinite metric cone (2-dimensional) with cone angle $\pi$ and cone point excised. It is this positive curvature which gives rise to the counterexample presented.

More generally, \ref{sec:meanonly} showed us that in these mean-only models there is a diagonal action on a subspace isometric to $(\R^{K-1})^d$. In all of these cases, under the action of $S_K$, the solid angle measure of a sphere about the origin will be divided by $K!$ when quotiented, producing a point of positive curvature, and leading to highly symmetric counterexamples with non-uniqueness of barycenters.

\end{document}